\documentclass[10pt,twocolumn,letterpaper]{article}

\usepackage[pagenumbers]{iccv} 

\usepackage{xcolor} 
\definecolor{googlered}{HTML}{DB4437} 
\definecolor{googleblue}{HTML}{4285F4} 
\definecolor{googlegreen}{HTML}{0F9D58} 
\definecolor{googleyellow}{HTML}{F4B400} 
\definecolor{googlepurple}{HTML}{9C27B0} 
\definecolor{googleorange}{HTML}{F4A11B} 

\newcommand{\redText}[1]{{\color{googlered}#1}}
\newcommand{\blueText}[1]{{\color{googleblue}#1}}
\newcommand{\greenText}[1]{{\color{googlegreen}#1}}

\newcommand{\purpleText}[1]{{\color{googlepurple}#1}}
\newcommand{\orangeText}[1]{{\color{googleorange}#1}}

\newcommand{\red}[1]{{\color{red}#1}}

\definecolor{iccvblue}{rgb}{0.21,0.49,0.74}
\usepackage[pagebackref,breaklinks,colorlinks,allcolors=iccvblue]{hyperref}

\def\confName{ICCV}
\def\confYear{2025}

\title{Conceptrol: Concept Control of Zero-shot Personalized Image Generation}

\author{
    Qiyuan He \quad Angela Yao \\
    National University of Singapore \\
    {\tt\small qhe@u.nus.edu \quad ayao@comp.nus.edu.sg}
}

\begin{document}

\twocolumn[{%
    \renewcommand\twocolumn[1][]{#1}%
    \vspace{-3em}
    \maketitle
    \begin{center}
        \centering
        \includegraphics[width=1.0\textwidth]{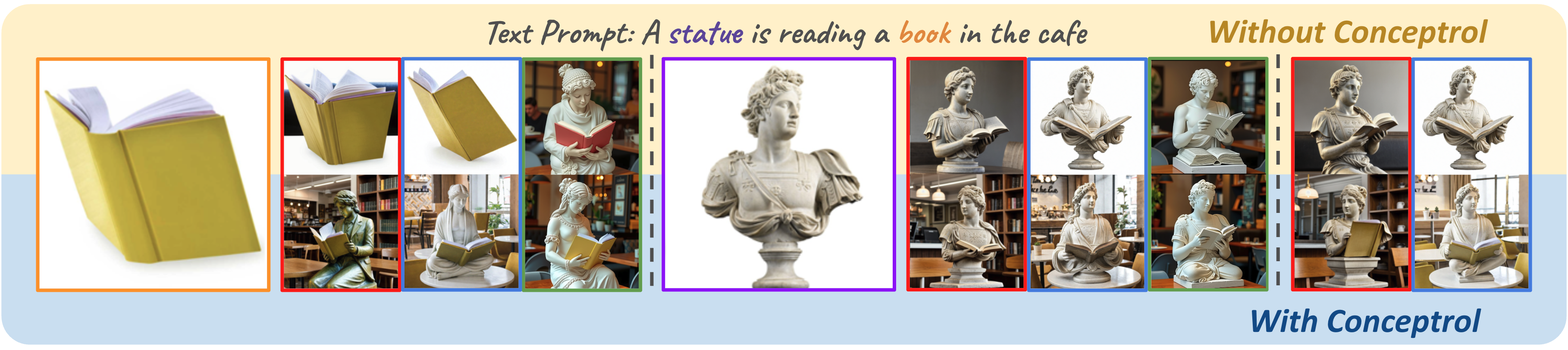}
        \begin{minipage}[t]{0.495\linewidth}
            \vspace{-10pt}
            \centering
            \includegraphics[width=\linewidth]{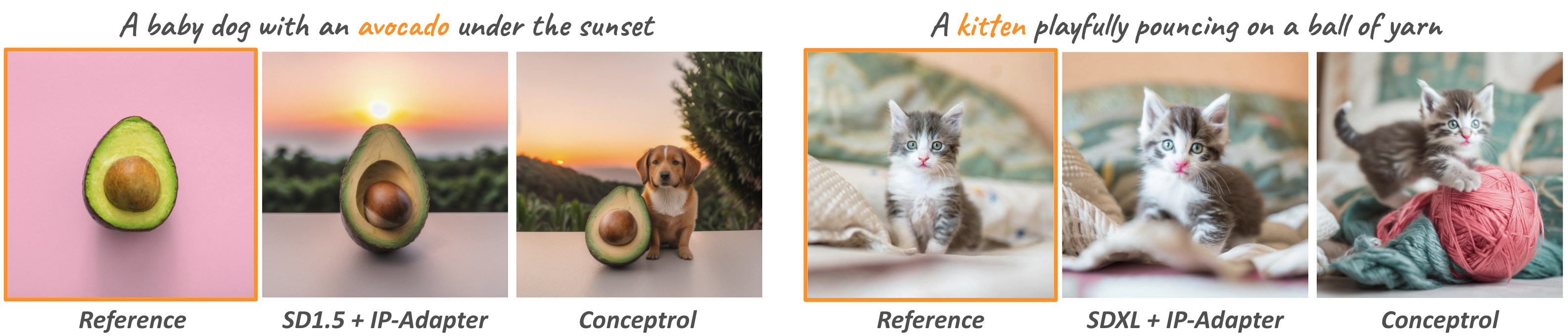}
        \end{minipage}%
        \hfill
        \begin{minipage}[t]{0.495\linewidth}
            \vspace{-10pt}
            \centering
            \includegraphics[width=\linewidth]{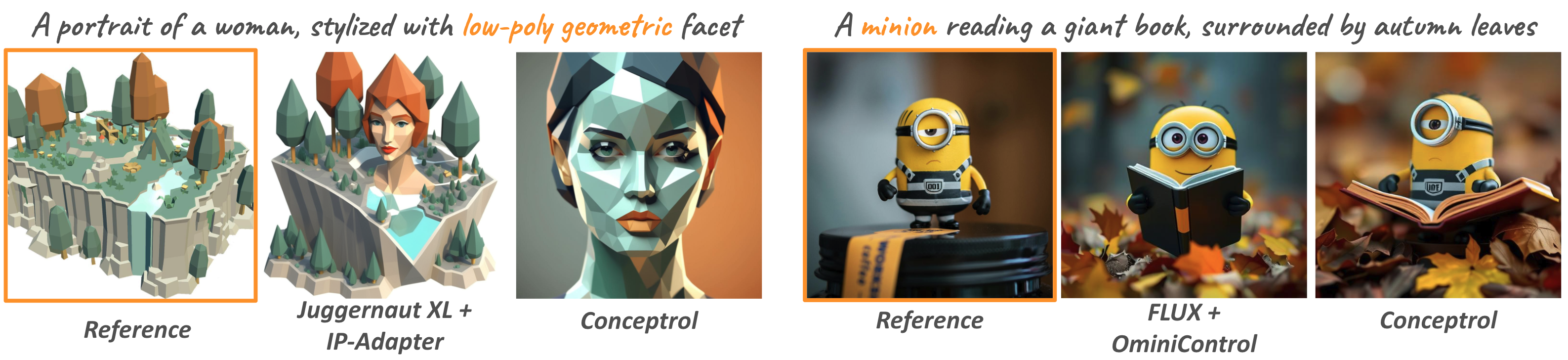}
        \end{minipage}
        \captionof{figure}{
        We propose \textit{\textbf{Conceptrol}}, a training-free control method that markedly improves the customization capabilities of zero-shot adapters. As the first row shows, adapters exhibit issues such as copy-paste artifacts (e.g., duplicating the book) and mismatched visual specifications (e.g., displaying a red book or inconsistent statute). In contrast, \textit{\textbf{Conceptrol}} accurately preserves the identity while strictly following the text prompt, and it can be applied to multiple personalized generations following the reference on \purpleText{statue} and \orangeText{yellow book} simultaneously. Our method supports different base models (\redText{Stable Diffusion}, \blueText{SDXL} and \greenText{FLUX}), personalized targets (e.g., objects and styles), and model parameters (e.g., SDXL and Juggernaut XL), all while without computation overhead, training data or auxiliary models.
        }
        \label{fig:teaser}
        \vspace{5pt}
    \end{center}%
    }]

\begin{abstract}
Personalized image generation with text-to-image diffusion models generates unseen images based on reference image content. Zero-shot adapter methods such as IP-Adapter and OminiControl are especially interesting because they do not require test-time fine-tuning.  However, they struggle to balance preserving personalized content and adherence to the text prompt. We identify a critical design flaw resulting in this performance gap: current adapters inadequately integrate personalization images with the textual descriptions. The generated images, therefore, replicate the personalized content rather than adhere to the text prompt instructions. 
Yet the base text-to-image has strong conceptual understanding capabilities that can be leveraged.

We propose Conceptrol, a simple yet effective framework that enhances zero-shot adapters without adding computational overhead. Conceptrol constrains the attention of visual specification with a textual concept mask that improves subject-driven generation capabilities. It achieves as much as 89\% improvement on personalization benchmarks over the vanilla IP-Adapter and can even outperform fine-tuning approaches such as Dreambooth LoRA.  The source code is available at \href{https://github.com/QY-H00/Conceptrol}{https://github.com/QY-H00/Conceptrol}.
\end{abstract}    
\section{Introduction}
\label{sec:intro}

Customizing or personalizing image generation with reference images is a significant and practical application of text-to-image diffusion models~\cite{rombach2022high, podell2023sdxl, chen2023pixart, flux2023, esser2024scaling}. 
Few-shot fine-tuning approaches like Textual Inversion~\cite{gal2022image} and DreamBooth~\cite{ruiz2023dreambooth} fine-tune an off-the-shelf diffusion model with a few reference images to yield a personalized model. However, as model sizes grow, fine-tuning becomes computationally expensive, limiting their efficiency for real-world applications. Another line of work focuses on training additional adapters, such as IP-Adapter~\cite{ye2023ip} or OminiControl~\cite{tan2024ominicontrol}, to enable zero-shot personalization. Adapters are more efficient during inference because there is no need for testing-time fine-tuning. However, they struggle to balance the preservation of the reference image concepts while adhering to the textual prompts. Recent research~\cite{tan2024ominicontrol,cai2024diffusion} attributes this challenge to a lack of training data that pairs the same reference subject with diverse textual descriptions. Yet even with additional data, zero-shot adapters are weaker in following textual instructions compared to base models without reference images~\cite{tan2024ominicontrol}.

This paper examines the weakness of zero-shot personalization from a design perspective. Our key insight is that fine-tuning approaches better leverage the textual concept behind the reference image. For example, given reference images of a specific cat, DreamBooth incorporates the textual concept of ``cat'' into the fine-tuning loss design. This allows them to exploit better the text understanding capabilities of the base diffusion model, such as Stable Diffusion. In contrast, zero-shot adapters inject the reference image as a condition in the attention block with Direct Adding~\cite{ye2023ip} or MM-Attention~\cite{flux2023, tan2024ominicontrol}. To that end, the reference image is not attached to any explicit text concept. This often results in a copy-paste effect~\cite{tan2024ominicontrol} or poor compositional generation, as demonstrated in Figure~\ref{fig:teaser}. 

Can we make adapters explicitly utilize textual concepts with minimal intervention and without re-training? To answer this question, we conduct {extensive analysis of the attention mechanism} quantitatively and qualitatively in Sec.~\ref{sec:attn_analysis}, as it is the primary interface to incorporate personalized reference images. Through the lens of the attention map, there are three main observations: 1) In the absence of textual concept constraints, the attention map for reference images derived from adapters does not focus on the target subject requiring customization; 2) Adapters do well at transferring appearance of reference images within regions of high attention scores; 3) There exists specific attention blocks in the text-to-image diffusion models, such as Stable Diffusion~\cite{rombach2022high}, SDXL~\cite{podell2023sdxl}, and FLUX~\cite{flux2023}, which consistently provide a~\textit{\textbf{textual concept mask}}, an attention map with strong attention on regions of customized target.

Building on insights (1) and (2), we consider that reference images are more effectively integrated as \textit{\textbf{visual specifications}} constrained by \textit{\textbf{textual concept mask}}. Attention masking~\cite{marcos2024open, zhou2024maskdiffusion, wu2023diffumask, wang2024compositional, endo2024masked} is well-established in pure text-to-image generation with manually provided regions of interest. We propose to apply masking on attention maps of the reference image, but \emph{without} any manual input.  Instead, based on insight (3), we exploit the inherent ability of the base diffusion models to identify the textual concept masks.

We introduce a simple but effective plug-and-play method called \textit{\textbf{Conceptrol}} to achieve this goal.  Conceptrol enhances the zero-shot customization of existing adapters in a training-free manner. During inference, given a global text prompt (e.g., ``a kitten playfully pouncing on a ball of yarn"), a manually provided textual concept (e.g., ``kitten"), and a visual specification (e.g., an image of a specific kitten) for customized target, we derive a textual concept mask from specific attention blocks in the base models, and use it to constrain the attention of the visual specification. Intuitively, our approach harnesses the original generation capabilities of text-to-image diffusion models and the appearance-transfer capabilities of lightweight adapters, improving the trade-off between concept preservation and prompt adherence.

We conduct comprehensive quantitative experiments using DreamBench++~\cite{peng2024dreambench++} and human evaluation from MTurk~\cite{mturk}.  We test Conceptrol with IP-Adapter~\cite{ye2023ip} for UNet-based diffusion models (e.g. Stable Diffusion, SDXL) and OminiControl~\cite{tan2024ominicontrol} for DiT-based models (e.g. FLUX). With Conceptrol, these zero-shot adapters make significant improvements by a large margin on personalized image generation benchmarks, even surpassing fine-tuning methods. To summarize, our contributions are as follows:

\begin{itemize}
    \item We identify a critical design flaw in zero-shot adapters, showing that neglecting textual concepts leads to incorrect attention in reference images.
    \item We discover that specific blocks in base models can provide a \textit{\textbf{textual concept mask}} that precisely indicates the spatial location of corresponding textual concepts.
    \item We introduce a simple yet effective method called \textit{\textbf{Conceptrol}}. We extract \textit{\textbf{textual concept mask}} and leverage it to increase the attention score of visual specification on the proper region of the personalized target while suppressing attention on irrelevant regions.
    \item Extensive evaluations show that Conceptrol improves zero-shot personalized image generation by a large margin and even surpasses fine-tuning methods despite its simplicity and negligible computational overhead.
\end{itemize}

Conceptrol highlights the importance of integrating textual concepts into personalized generation pipelines. The source code is available \href{https://github.com/QY-H00/Conceptrol}{here}.
\section{Related Work}

\noindent\textbf{Text-to-Image Diffusion Models} have achieved state-of-the-art image generation quality~\cite{rombach2022high, podell2023sdxl, flux2023, esser2024scaling} using text prompts and training on massive, diverse datasets like LAION-5B~\cite{schuhmann2022laion}. Previous models like Imagen~\cite{saharia2022photorealistic}, Stable Diffusion 1.5~\cite{rombach2022high}, and SDXL~\cite{podell2023sdxl} used a UNet architecture~\cite{ronneberger2015u}. More recently, the Diffusion Transformer (DiT)~\cite{peebles2023scalable} with flow matching~\cite{lipman2022flow} has become the dominant design in models such as Stable Diffusion 3~\cite{esser2024scaling} and FLUX~\cite{flux2023}. Both kinds of models integrate text conditions with the noisy latent through attention mechanisms. Empirically, they demonstrate strong text comprehension capabilities that can be leveraged for personalized generation.

\noindent\textbf{Personalized Image Generation} creates images from text prompts and reference images that depict previously unseen, customized targets~\cite{gal2022image, ruiz2023dreambooth, ye2023ip, tan2024ominicontrol}. A key challenge in this domain is preserving the novel concept while ensuring consistency with the text prompt, as highlighted by the subject-driven generation benchmark DreamBenchPlus~\cite{peng2024dreambench++}. This challenge can be framed as a multi-task problem, requiring concept preservation and prompt adherence, and is analogous to maximizing a Nash welfare~\cite{nash1950bargaining}.

\noindent\textbf{Fine-tuning Methods} adapt diffusion models given personalized content for further generation. Textual Inversion~\cite{gal2022image} freezes all model parameters and tunes only the text embedding associated with reference images. Dreambooth~\cite{ruiz2023dreambooth} fine-tunes the entire diffusion model while regularizing against the hyper-class of the target to leverage the base model’s semantic prior. While effective, these methods are computationally expensive. Also, the need to fine-tune a separate model for each target limits scalability, motivating the development of zero-shot adapters.

\noindent\textbf{Zero-shot Adapters} like IP-Adapter~\cite{ye2023ip} and OminiControl~\cite{tan2024ominicontrol} bypass fine-tuning by training additional components to incorporate image conditions directly into large-scale diffusion models. IP-Adapter~\cite{ye2023ip} uses CLIP image embeddings and cross-attention layers to integrate image conditions alongside text prompts for UNet-based models. OminiControl~\cite{tan2024ominicontrol} is an adapter for DiT-based models like FLUX~\cite{flux2023}. It appends image condition tokens to the multi-modal attention layers and trains a LoRA adapter for efficient adaptation. These methods reduce the computational overhead of inference but their performance lags behind fine-tuning methods, even with large amounts of training data~\cite{ye2023ip, tan2024ominicontrol}. This suggests there may be fundamental design limitations to the current zero-shot adapters.

\noindent\textbf{Controlled generation} use masked attention to restrict the generation within a given region~\cite{marcos2024open, zhou2024maskdiffusion, wu2023diffumask, wang2024compositional, endo2024masked}. ~\citet{endo2024masked} proposed to integrate user-defined masks, enabling precise spatial control without additional training. Similarly,~\citet{wang2024compositional} developed an attention mask control strategy using predicted bounding boxes to guide attention regions. These approaches demonstrate that masking in the attention layer can effectively limit the generation region.
\begin{figure}
    \centering
    \includegraphics[width=\linewidth]{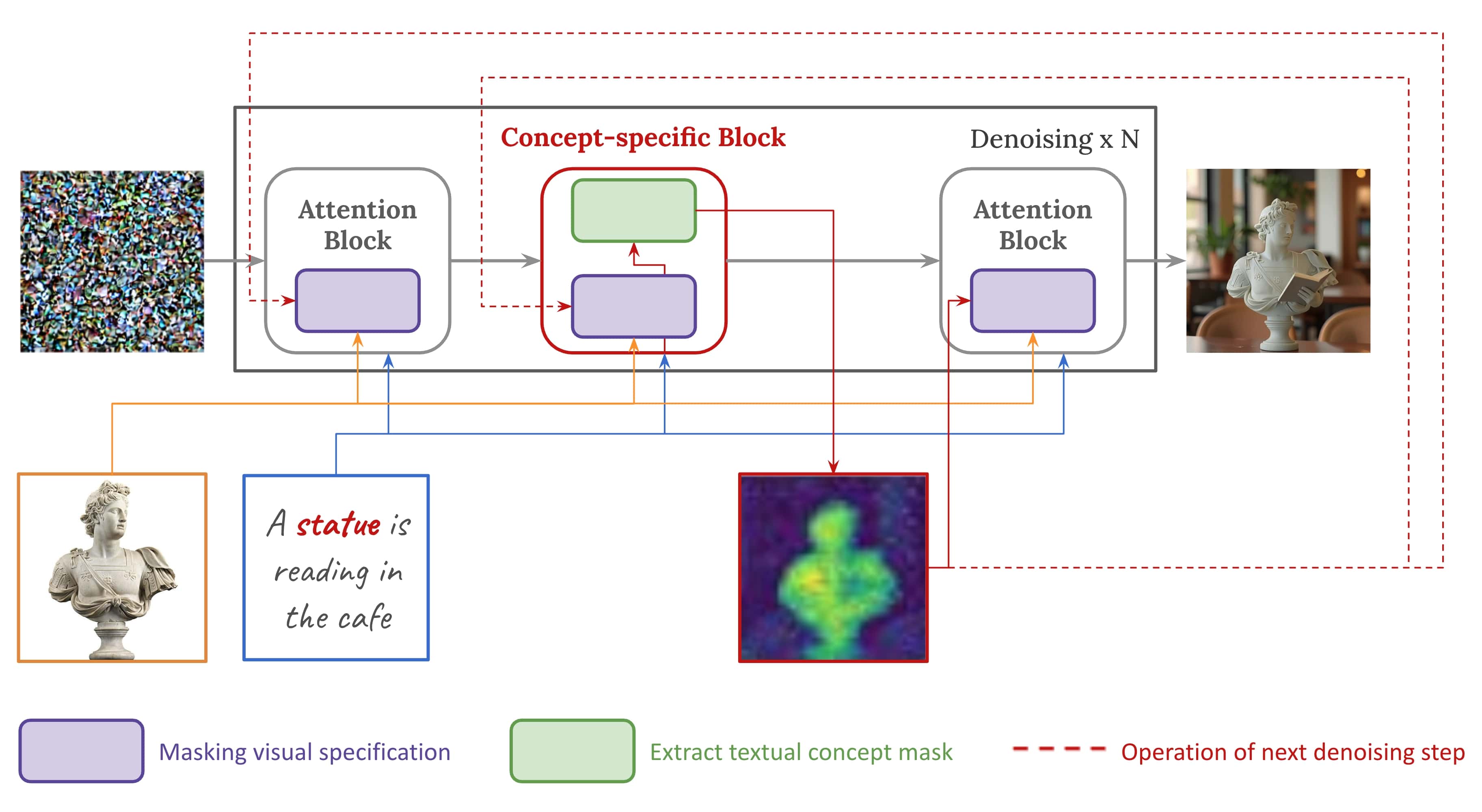}
    \caption{\textbf{Overview of Conceptrol}. Conceptrol extracts a textual concept mask indicating the region of interest for a textual concept (e.g., "statue"), from a concept-specific block (e.g., \texttt{UP BLOCK 0.1.3} in SDXL). It then adjusts the attention to the corresponding visual specification (i.e., a personalized image of the statue) in the adapters accordingly to enhance personalization.}
    \label{fig:method_pipe}
\end{figure}

\section{Methodology}

\subsection{Preliminaries}
\label{sec:preliminaries}
\noindent\textbf{Text-to-Image Diffusion Models} generate images by progressively denoising a latent, $x_T$, sampled from a standard Gaussian distribution. The process is guided by a denoising function, $\epsilon_\theta$, and conditioned on a text prompt, $c_{\text{text}}$. The reverse diffusion process conditionally samples latent $x_{t-1}$, in a latent space encoded by a VAE~\cite{kingma2013auto}:

\begin{equation}\label{eq:dif}
p(x_{t-1} | x_t, c_{\text{text}}) = \mathcal{N}(x_{t-1}; \mu_\theta(x_t, t, c_{\text{text}}), \Sigma_\theta(t)).
\end{equation}

The mean in~\cref{eq:dif}, $\mu_\theta(x_t, t, c_{\text{text}})$, is computed as:

\begin{equation}
\mu_\theta(x_t, t, c_{\text{text}}) = \frac{1}{\sqrt{\alpha_t}} \left( x_t - \frac{\beta_t}{\sqrt{1 - \bar{\alpha}_t}} \epsilon_\theta(x_t, t, c_{\text{text}}) \right),
\end{equation}

\noindent with hyperparameters $\alpha_t$ and $\beta_t$,  while $\Sigma_\theta(t)$ represents a fixed or learned variance. The denoising function $\epsilon_\theta$ is represented by a UNet~\cite{ronneberger2015u, rombach2022high, podell2023sdxl} or a ViT~\cite{dosovitskiy2020image, flux2023}. Starting with $x_T$, the denoising process iteratively refines the latent to reach $x_0$, which is then decoded into an image with a VAE decoder. Note that FLUX~\cite{flux2023} uses flow matching~\cite{lipman2022flow} with a standard Gaussian distribution for generative modeling, which is also considered as a form of diffusion~\cite{wang2024rectified}.

\noindent\textbf{Notation used in Conceptrol}. Let \( c_{\text{image}} \) denote the \textit{visual specification}, which corresponds to the provided reference image used for personalization. Let \( c_{\text{text}} \) denote the \textit{text condition}, a representation corresponding to the text description of the desired output. Formally, \( c_{\text{text}} \) is represented as \( c_{\text{text}} \in \mathbb{R}^{N \times C} \), where \( N \) is the number of tokens in the text and \( C \) is the dimensionality of each token. We further introduce the notion of a \textit{textual concept}, \( c_{\text{concept}} \), defined as a substring of \( c_{\text{text}} \). For instance, in the prompt \textit{``A dog is playing with a cute cat in the living room,''} the textual concept can be \textit{``a cute cat''} if the visual specification provided represents a cat. The textual concept \( c_{\text{concept}} \) is sliced from $c_{\text{text}}$ as $c_{\text{concept}} = c_{\text{text}}[i_s:i_e, :] \in \mathbb{R}^{N^\prime \times C}$, where \([i_s, i_e]\) are the start and end indices (excluded) of the textual concept within \( c_{\text{text}} \). Here, \( N^\prime = i_e - i_s \) denotes the number of tokens representing the textual concept.

\noindent\textbf{IP-Adapter for U-Net Models: Direct Adding.} UNet models~\cite{saharia2022photorealistic, podell2023sdxl} incorporate conditions using cross-attention. For a noisy latent $x_t$, a condition $c$, and matrices $W_q^{(l)}$, $W_k^{(l)}$, and $W_v^{(l)}$ of attention block $l$, the cross-attention is computed as:

\begin{equation}\label{eq:xattn}
\text{Attn}^{(l)}(x_t, c) = \underbrace{\text{Softmax}\left(\frac{(W_q^{(l)} x_t)(W_k^{(l)} c)^T}{\sqrt{d}}\right)}_{A^{(l, x_t)}_c} W_v^{(l)} c,
\end{equation}

\noindent where $d$ is the dimensionality of the query and key vectors, ${A^{(l, x_t)}_c}$ is the attention map. In text-to-image generation, UNet models apply~\cref{eq:xattn} with a text condition $c_{\text{text}}$ as $c$. The IP-Adapter extends this mechanism by introducing an additional cross-attention for the reference image represented by CLIP embedding, $c_{\text{image}}$. The combined attention $\text{Attn}_{\text{IP}}^{(l)}$ sums the cross attention of $c_{\text{text}}$ and 
$c_{\text{image}}$:

\begin{equation}
\begin{aligned}
&\text{Attn}_{\text{IP}}^{(l)}(x_t, c_{\text{text}}, c_{\text{image}}; \lambda) \\ = &\text{Attn}^{(l)}(x_t, c_{\text{text}}) + \lambda \cdot \text{Attn}^{(l)}(x_t, c_{\text{image}}).
\end{aligned}
\end{equation}

\noindent Above, $\lambda$ represents the \textit{IP Scale}, a weighting hyperparameter on the reference image.

\noindent\textbf{OminiControl for DiT Models: MM-Attention.} DiT models use multi-modal attention to fuse text conditions to the latent. OminiControl further incorporates an image condition in the fused token. Given a noisy latent $x_t$, a text condition $c_{\text{text}}$ and an image condition $c_{\text{image}}$, the fused latent is represented as $x^\prime_t = [c_{\text{text}}, x_t, c_{\text{image}}]$, where $[\cdot, \cdot, \cdot]$ denotes token concatenation. Then the attention is computed as:

\begin{equation}
\begin{aligned}
&\text{Attn}^{(l)}_{\text{omini}}(x_t, c_{\text{text}}, c_{\text{image}}; \lambda) \\ = &\underbrace{\text{Softmax}\left(\frac{(W_q^{(l)} x^\prime_t)(W_k^{(l)} x^\prime_t)^T}{\sqrt{d}} + B(\lambda)\right)}_{A_{mm}^{(l, x_t)}} W_v^{(l)} x^\prime_t.
\end{aligned}
\label{eq:omini_formula}
\end{equation}

\noindent where $B(\lambda)$ is used to control the conditioning scale of image and $A_{mm}^{(l, x_t)}$ is the fused attention map. Given $c_{\text{text}} \in R^{M \times d}$ , $x_t, c_{\text{image}} \in R^{N \times d}$, it is computed as:

\begin{equation}
B(\lambda) =
\begin{bmatrix}
\mathbf{0}_{M \times M} & \mathbf{0}_{M \times N} & \mathbf{0}_{M \times N} \\
\mathbf{0}_{N \times M} & \mathbf{0}_{N \times N} & \log(\lambda) \mathbf{1}_{N \times N} \\
\mathbf{0}_{N \times M} & \log(\lambda) \mathbf{1}_{N \times N} & \mathbf{0}_{N \times N}
\end{bmatrix}.
\label{eq:omini_lambda}
\end{equation}

In Eqs.~\ref{eq:omini_formula} and ~\ref{eq:omini_lambda}, a larger $\lambda$ increases the scaling of the image conditioning. Compared to IP-Adapter, OminiControl is trained using a larger dataset of data pairs featuring the same subjects with diverse text prompts.

\begin{figure}
    \centering
    \includegraphics[width=\linewidth]{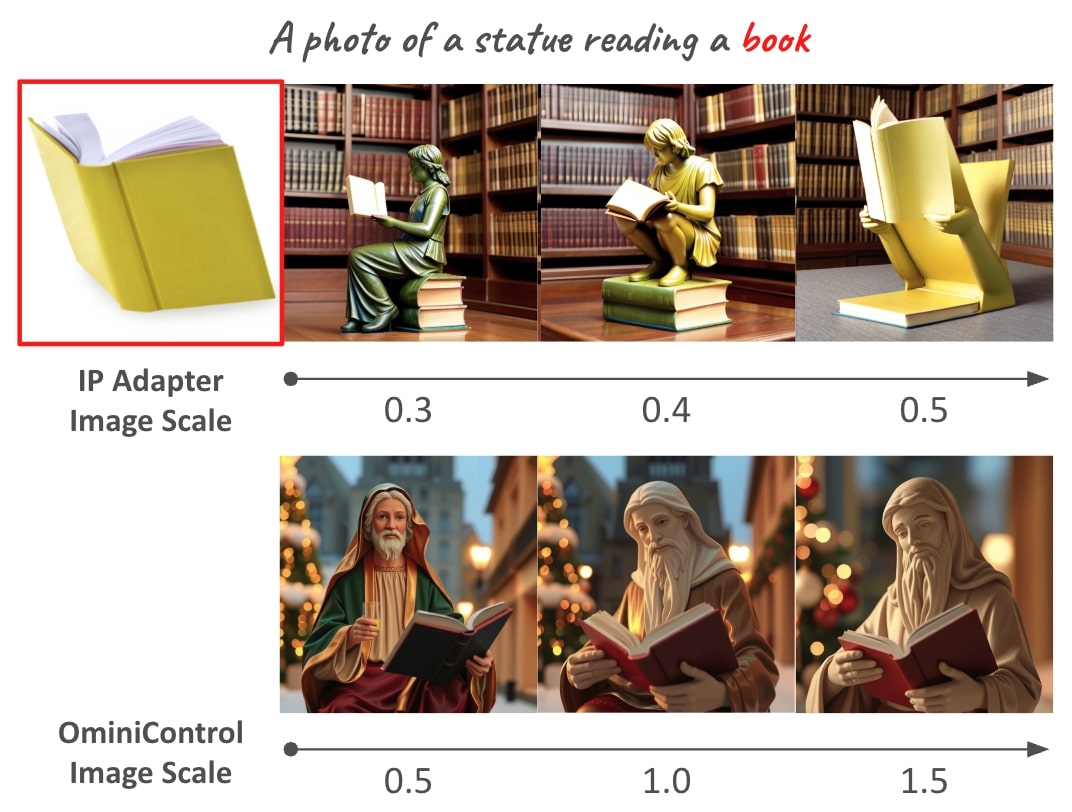}
    \caption{\textbf{Treating image conditions globally can be problematic.} The first row shows IP-Adapter on Stable Diffusion 1.5 with varying IP Scales, where increasing scale shifts the output from "a statue reading a book" to "a book statue." The second row shows OminiControl on FLUX failing to preserve the color of the book as yellow but generating red books at different conditioning scales.}
    \label{fig:cp_pf_imbalance}
\end{figure} 

\subsection{Why is it suboptimal to treat the reference image as a global condition?}

For both IP-Adapter and OminiControl, the reference image is used as a global condition based on two observations. Firstly, during training, the image condition always serves as the main subject presented in the target of generation~\cite{ye2023ip, tan2024ominicontrol}. Furthermore, the image condition and text condition in these adapters are mathematically symmetric in the formulation, i.e., exchanging $c_{\text{text}}$ and $c_{\text{image}}$ will not change the modeling, where text condition already serves as a global description of the generated content.

Two main challenges of zero-shot personalization stem from treating the image and text conditions symmetrically: 1) balancing prompt adherence with concept preservation and 2) obtaining diverse datasets of the same subject described with varying prompts. First, while text conditions serve as global prompts describing the desired image, treating the image conditions similarly can lead to conflicts. As shown in Fig.~\ref{fig:cp_pf_imbalance} (row 1), a low image conditioning strength (IP Scale) in IP-Adapter~\cite{ye2023ip} fails to preserve the concept effectively while increasing the scale causes deviations from the text prompt and leads to a copy-paste effect. This highlights the difficulty in balancing prompt adherence with concept preservation if the conditions are treated similarly.

Second, the coupling between text and image conditions presents significant challenges. This holds even when training with data pairs of the same subject with different text prompts, as done in OminiControl~\cite{tan2024ominicontrol},  For example, when generating ``A statue is reading a book'' with reference images of a specific book, the system may mistakenly prioritize ``statue'' over ``book'' and simply ignore the reference image of the book, as shown in Fig.~\ref{fig:cp_pf_imbalance} (row 2).

Instead of treating the image and text conditions equally, we propose to use the image condition as a visual specification of a particular textual concept. For instance, in the prompt "A photo of a statue reading a book", the image condition should only be applied to the generation of ``a book'' rather than the whole scene. Otherwise, as Fig.~\ref{fig:cp_pf_imbalance} (row 1) shows, the reference might affect the generation of ``statue'' as well and lead to artifacts.  

\begin{figure}
    \centering
    \includegraphics[width=\linewidth]{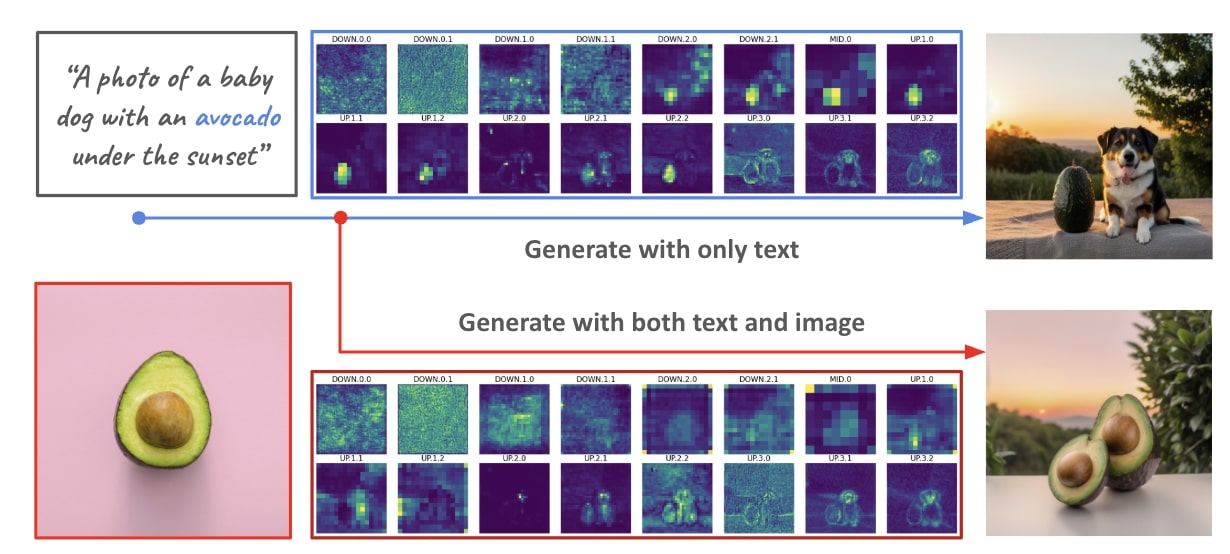}
    \caption{\textbf{Incorrect attention map of image conditions.} This example illustrates IP-Adapter results with {\color{blue}fully text-based input} and {\color{red}with additional image condition} added at the 10th of 50 total denoising steps. The blue box shows the attention map of 'avocado,' while the red box highlights the image condition, which incorrectly focuses on the dog area as well with the given avocado image, distorting results and reducing text prompt adherence.}
    \label{fig:attn_analysis}
\end{figure}

\subsection{What does the attention of noisy latent to the text and image conditions indicate?}
\label{sec:attn_analysis}

Since the attention block is the primary mechanism through which both IP-Adapter~\cite{ye2023ip} and OminiControl~\cite{tan2024ominicontrol} incorporate additional image conditions, we investigate how these conditions interact with the noisy latent representations and influence the generation process. Prior works have analyzed attention maps post-hoc—after the full generation~\cite{marcos2024open, zhou2024maskdiffusion}; however, they focus solely on text-to-image generation.  Our analysis differs in two respects. First, we analyze attention maps to determine whether the region of interest can be identified \textit{without prior knowledge during generation}. Second, we examine how additional reference images influence the generation process.

\begin{figure}[t]
    \centering
    \includegraphics[width=\linewidth]{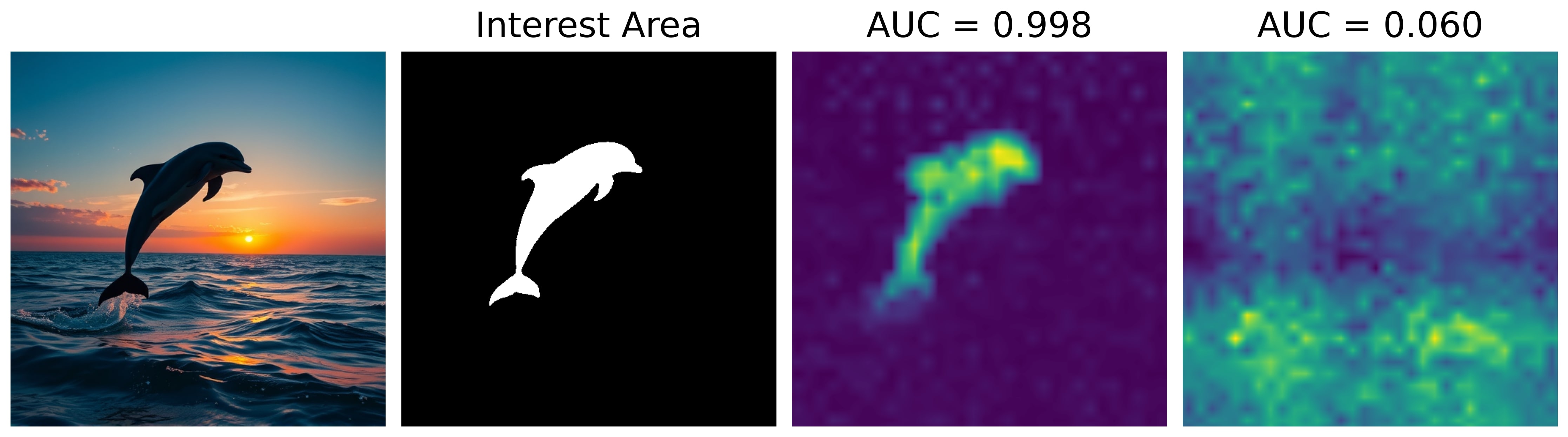}
    \begin{minipage}[b]{0.25\linewidth}
        \centering \small (a)
    \end{minipage}%
    \begin{minipage}[b]{0.25\linewidth}
        \centering \small (b)
    \end{minipage}%
    \begin{minipage}[b]{0.25\linewidth}
        \centering  \small (c)
    \end{minipage}%
    \begin{minipage}[b]{0.25\linewidth}
        \centering  \small (d)
    \end{minipage}%
    \caption{\textbf{Not all attention maps of textual concept strongly indicate the interest area.} Shown are examples from the FLUX model including (a) generated results for ``dolphin'', (b) segmentation results from SAM indicating the subject, and attention map with \textit{"dolphin"} from (c) \texttt{BLOCK 18} and (d) \texttt{BLOCK 11}. }
    \label{fig:auc_attn_mask}
\end{figure}

To explore these questions, we first analyze the generation process without reference images by setting the conditioning scale to zero, while computing the attention map for the reference image. Next, we use LangSAM~\cite{medeiros2024lang}, an open-vocabulary segmentation tool based on SAM~\cite{kirillov2023segment}, to obtain pseudo masks of the customization target. For instance, in Fig.~\ref{fig:auc_attn_mask}, (b) shows the mask produced by LangSAM, while (c) displays one of the attention maps. By calculating the AUC between the attention map and the target mask, we quantitatively assess whether the attention map correctly highlights the target’s region of interest. We provide more details of our analysis in the Appendix.~\red{A}.

\noindent\textbf{Attention Distribution for Image Conditions is Misaligned.} Fig.~\ref{fig:attn_analysis} shows an example of this discrepancy. In this example, the attention map corresponding to the text “avocado” closely matches the ground-truth mask of the avocado in the generated result, whereas the attention map of the image condition is focused on unrelated objects such as the dog. Quantitatively, the highest AUC of the image condition's attention maps across all blocks is only 0.38, compared to an AUC of up to 0.99 for the text (e.g., “avocado”). This supports our insights in Sec.~\ref{sec:intro} that the attention maps from reference images do not focus on the target subject requiring customization.

\noindent\textbf{Visual specifications can be transferred within regions of high attention score}. It has been shown that IP-Adapter can transfer visual specification by manually applying attention mask~\cite{ye2023ip}. We further verify this in FLUX with OminiControl as well. Specifically, we use the mask indicating the region-of-interest segmented from the results by text-only conditions, and then use it to mask image-condition generation to get another result. Then we segment from the new result again, and compare with the original masking, The AUC can be as high as \textbf{0.99} for both UNet-based model and DiT-based models. This indicates that adapters can transfer appearance of reference images within regions of high attention scores.

\begin{figure}
    \centering
    \begin{minipage}[b]{0.3\linewidth}
        \centering
        \includegraphics[width=\linewidth]{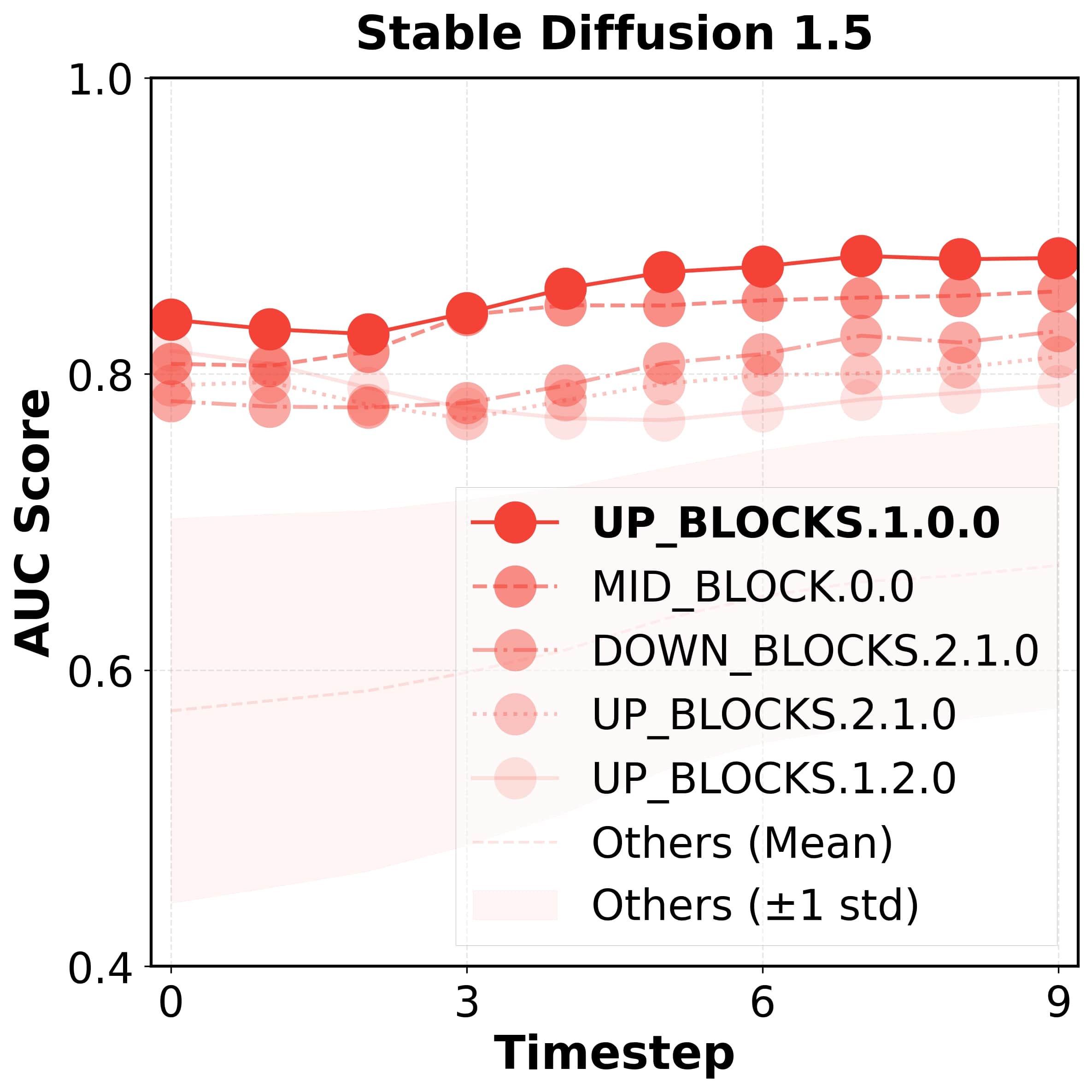} 
    \end{minipage}
    \hfill
    \begin{minipage}[b]{0.3\linewidth}
        \centering
        \includegraphics[width=\linewidth]{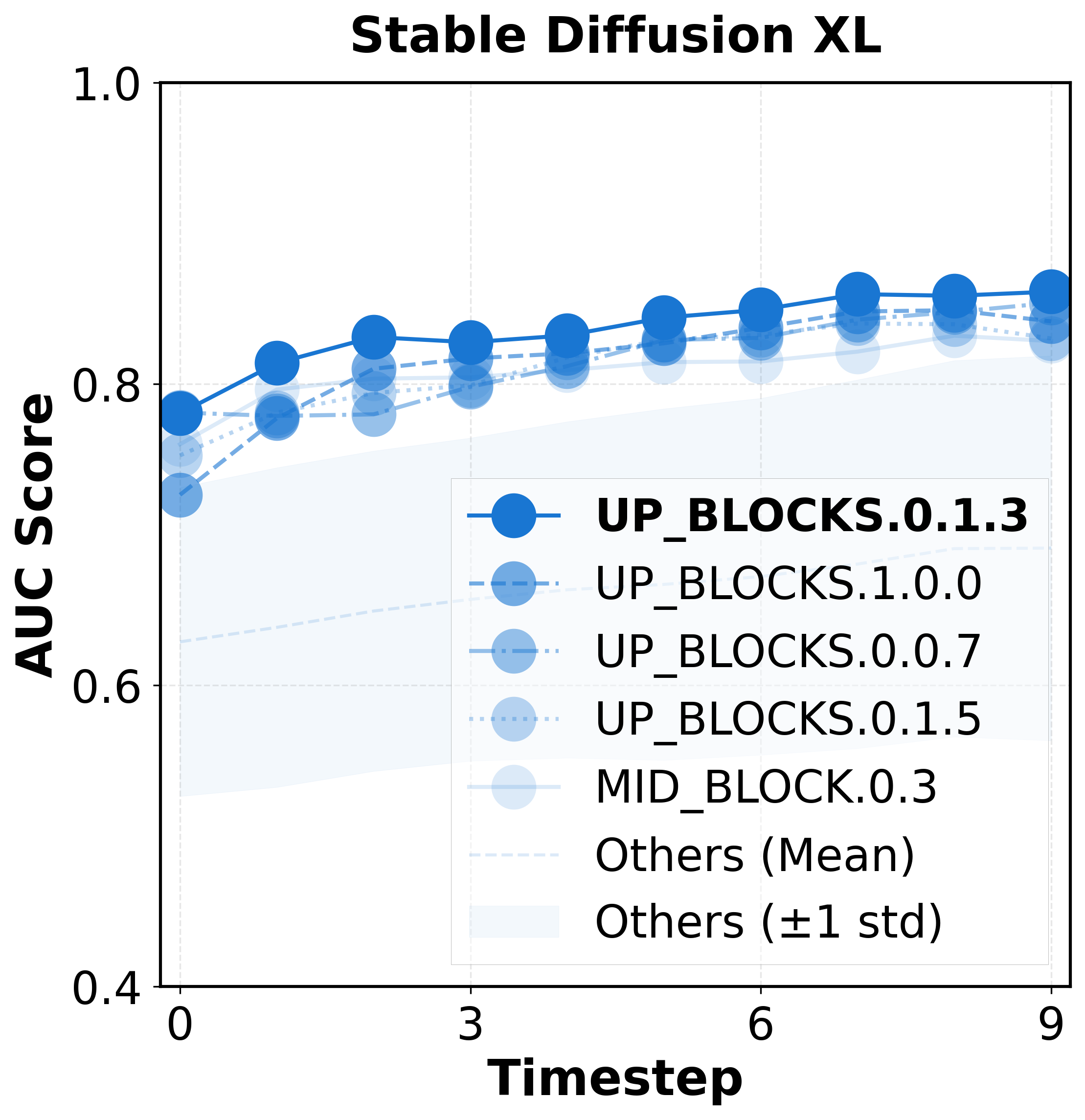}
    \end{minipage}
    \hfill
    \begin{minipage}[b]{0.3\linewidth}
        \centering
        \includegraphics[width=\linewidth]{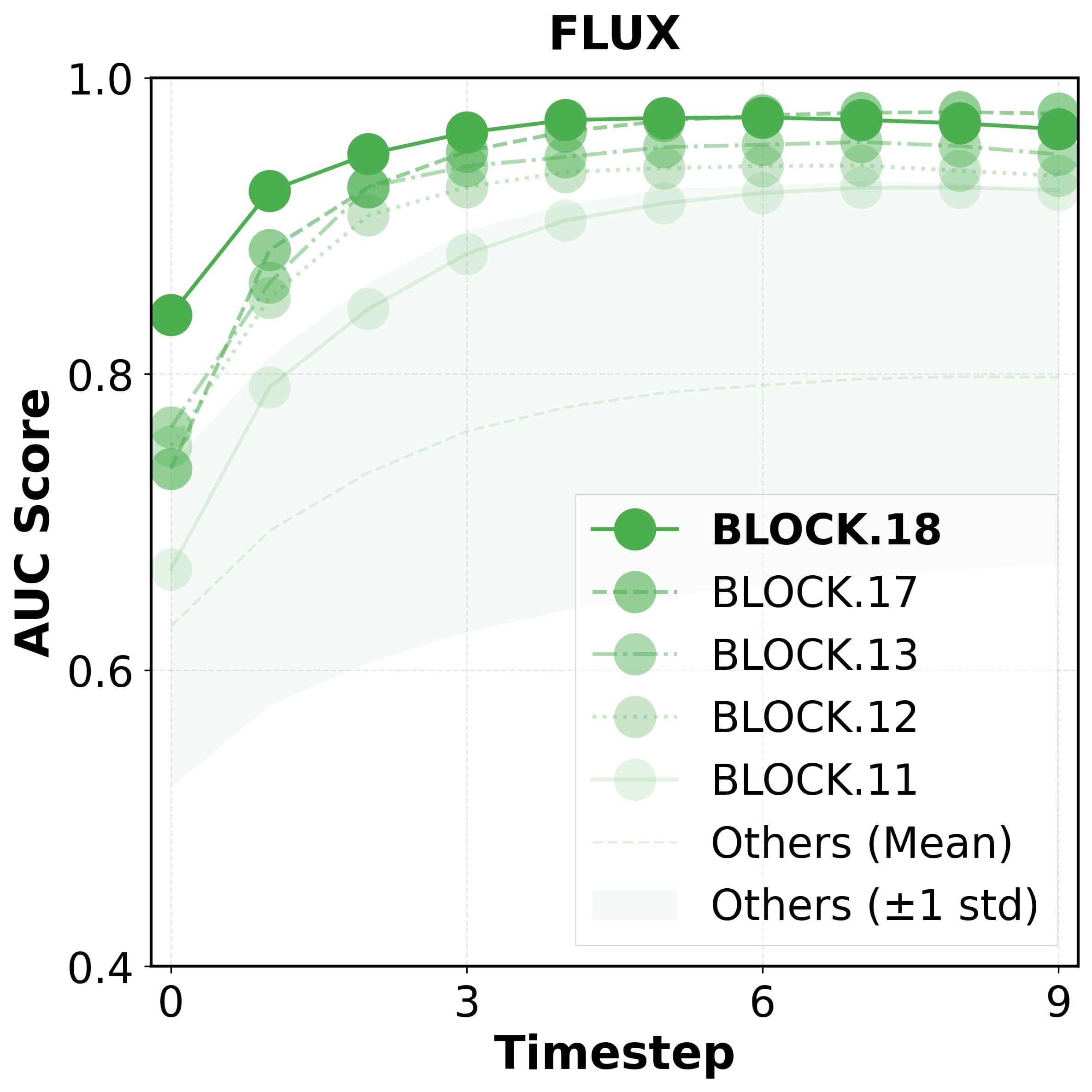} 
    \end{minipage}
    \caption{\textbf{The attention map of a textual concept directly reveals the regions of interest across various architectures.} We present the AUC scores of the attention maps for the textual concept from different blocks, compared to the ground truth mask, across timesteps in Stable Diffusion 1.5, SDXL, and FLUX. We can extract \textbf{\textit{textual concept mask}} from the \textbf{concept-specific block} alongside the generation.}
    \label{fig:concept_specific_block}
\end{figure}

\noindent\textbf{Concept-specific attention blocks for text conditions indicate the region of interest during generation.} Unlike previous post-hoc analyses of attention maps~\cite{marcos2024open}, we investigate their characteristics \emph{during} the generation process. For architectures that incorporate cross-attention or multi-modal attention to introduce text conditions, we observe specific blocks that clearly highlight the region of interest as shown in Fig.~\ref{fig:auc_attn_mask}. To quantify this, we compare the AUC between attention maps at each block and timestep with the annotated region of interest obtained via LangSAM. As illustrated in Fig.~\ref{fig:concept_specific_block}, attention maps from the concept-specific block strongly indicate the region of interest. These concept-specific blocks include the following: \texttt{UP BLOCK 1.0.0} in Stable Diffusion, \texttt{UP BLOCK 0.1.3} in SDXL, and \texttt{BLOCK 18} in FLUX. We refer to the maps from these specific blocks as \textbf{\textit{textual concept masks}}, as they directly correspond to the textual concept in the pixel space. Our third insight gained from the analysis is that: these specific blocks can consistently provide attention maps with high scores on regions of customized target.

\section{Conceptrol: Controlling visual specifications with a textual concept mask}

We propose a simple yet effective method to consistently boost the personalization ability of the zero-shot adapter, which we refer to as \textit{\textbf{Conceptrol}}. Building on the insights, Conceptrol employs a \textit{\textbf{textual concept mask}} to adjust the attention map of image conditions so that the region of personalized target receives the highest score, enabling the adapters to accurately transfer the \textit{\textbf{visual specification}}. The overall pipeline is demonstrated in Fig.~\ref{fig:method_pipe}.

\subsection{Conceptrol on Direct Adding / IP-Adapter}

For the concept-specific attention block $l^*$, such as \texttt{UP BLOCK 0.1.3} in SDXL, its attention map at inference timestep $t$ is obtained as $A^{(l^*, x_t)}_{c_\text{text}} \in R^{h \times (H \times W) \times M}$, where $h$ is the number of heads, $H, W$ is the size of the feature map and $M$ is the number of text tokens. We slice out the attention map of textual concept $A^{(l^*, x_t)}_{c_\text{concept}} = A^{(l^*, x_t)}_{c_\text{text}}[:, :, i_s:i_e]$. $A^{(l^*, x_t)}_{c_\text{concept}}$ is averaged over the heads and textual concept tokens and normalized to obtain $\widetilde{A}^{(l^*, x_t)}_{c_\text{concept}} \in R^{(H \times W)}$:

\begin{equation}
\begin{aligned}
\widetilde{A}^{(l^*,x_t)}_{c_\text{concept}} &= \text{mean}(A^{(l^*, x_t)}_{c_\text{concept}}, \text{dim}= \{0, 2\}), \\
\widetilde{A}^{(l^*, x_t)}_{c_\text{concept}} &:= \frac{\widetilde{A}^{(l^*, x_t)}_{c_\text{concept}}}{\text{max}(\widetilde{A}^{(l^*, x_t)}_{c_\text{concept}})}
\end{aligned}
\label{eq:normalize_mask}
\end{equation}

During inference, the IP-Adapter's cross-attention can be modified by masking the attention with the image condition:

\begin{equation}
\begin{aligned}
&\text{Attn}^{(l)}_{\text{Conceptrol+IP}}(x_t, c_{\text{text}}, c_{\text{image}}, \widetilde{A}^{(l^*, x_t)}_{c_\text{concept}}) \\ = &\text{Attn}^{(l)}(x_t, c_{\text{text}}) + \lambda 
\cdot \widetilde{A}^{(l^*, x_t)}_{c_\text{concept}} \odot \text{Attn}^{(l)}(x_t, c_{\text{image}})
\end{aligned}
\end{equation}

\noindent Where $\odot$ corresponds to element-wise multiplication on the spatial feature.

\subsection{Conceptrol on MM-Attention / OminiControl}

Similar to Conceptrol on Direct Adding, given concept-specific attention block $l^*$ (\texttt{BLOCK 18} in FLUX), we compute its attention map firstly as $A_{mm}^{(l^*, x_t)} \in R^{(h \times (M+2N) \times (M+2N))}$, where $h$ is the number of heads, $M$ is the number of text tokens, $N$ is the number of latent tokens, at inference timestep $t$ with concatenated tokens $[c_{\text{text}}, x_t, c_{\text{image}}]$. Based on this attention map, we can slice the attention map of noisy latent with textual concept by:$A_{c_\text{concept}}^{(l^*, x_t)} = A_{\text{mm}}^{(l^*, x_t)}[:,M:M+N, i_s:i_e]$. Similar to Eq.~\ref{eq:normalize_mask}, we average over each head and text tokens in textual concept $\widetilde{A}^{(l^*, x_t)}_{c_\text{concept}}$ but normalizing it by its mean value.

Additionally, different from Direct Adding, MM-Attention enforces attention between text and image condition as well. To further constraint influence of the image condition on irrelevant concepts, we define mask $M^\prime(\lambda
) \in R^{(M \times N)}$ as:
\vspace{-1px}
\begin{equation}
    M^\prime_{ij}(\lambda) = \log(\epsilon) + \mathbf{1}_{\{i_s \leq i < i_e\}} \left[ \log(\lambda) - \log(\epsilon) \right]
\end{equation}

\noindent Here $\lambda$ is the conditioning scale of reference images and $\epsilon$ is a value close to 0 such that $\log(\epsilon)$ is extremely small to prevent irrelevant text tokens paying attention to the reference image. Then we alter the attention in OminiControl as:
\begin{equation}
    \begin{aligned}
&\text{Attn}_{\text{Conceptrol+Omini}}(x_t, c_{\text{text}}, c_{\text{image}}, \widetilde{A}^{(l^*, t)}_{c_\text{concept}}) \\ = &\text{Softmax}\left(\frac{(W_q x^\prime_t)(W_k x^\prime_t)^T}{\sqrt{d}} + B_{c_\text{concept}}(\lambda)\right) W_v x^\prime_t.
\end{aligned}
\end{equation}

\noindent where $B_{c_\text{concept}}(\lambda)$ is computed as:

\begin{equation}
\scalebox{0.75}{$
B_{c_\text{concept}}(\lambda) =
\begin{bmatrix}
\mathbf{0}_{M \times M} & \mathbf{0}_{M \times N} & M^\prime(\lambda) \\
\mathbf{0}_{N \times M} & \mathbf{0}_{N \times N} & \log\left(\lambda{\widetilde{A}^{(l^*, x_t)}_{c_\text{concept}}}\right) \cdot \mathbf{1}_{1 \times N} \\
\mathbf{0}_{N \times M} & \log(\lambda) \mathbf{1}_{N \times N} & \mathbf{0}_{N \times N}
\end{bmatrix}
$}
\end{equation}

\noindent Here '$\cdot$' represents matrix multiplication. We provide more details and visualization in the Appendix.~\red{C}.

\subsection{Conceptrol Warmup}

We notice that the attention map is less informative in the early stage to indicate the region of interest as shown in Fig.~\ref{fig:attn_analysis} and Fig.~\ref{fig:concept_specific_block}. Therefore, instead of starting control initially, we introduce another hyperparameter, \textbf{conditioning warmup ratio} $\epsilon$, to prohibit the injection of image condition before a preset time step $T^\prime = \epsilon T$, where $T$ is the total inference timesteps. At each timestep $t > T^\prime$, for those block ahead of $l^*$, we use $\widetilde{A}^{(l^*, x_{t-1})}_{c_\text{concept}}$ accordingly. Otherwise, we use $\widetilde{A}^{(l^*, x_t)}_{c_\text{concept}}$ to control the visual specification. More details can be found in the Appendix.~\red{C}.

\section{Experiments}

\begin{table*}[t]
\centering
\scalebox{0.77}{
\begin{tabular}{c|ccccc|cccc|c}
\toprule
Method & \multicolumn{5}{c|}{Concept Preservation (CP)} & \multicolumn{4}{c|}{Prompt Following (PF)} & CP $\cdot$ PF \\
 & Animal & Human & Object & Style & Overall & Photorealistic & Style Transfer & Imaginative & Overall & \textbf{Final Score} \\
\midrule
Textual Inversion SD & 0.501 & 0.372 & 0.308 & 0.358 & 0.381 & 0.679 & 0.698 & 0.440 & 0.632 & 0.241 \\
DreamBooth SD & 0.646 & 0.196 & 0.491 & 0.476 & 0.496 & 0.789 & 0.778 & 0.510 & 0.723 & 0.359 \\
BLIP SD* & 0.676 & 0.556 & 0.469 & 0.508 & 0.548 & 0.578 & 0.518 & 0.300 & 0.496 & 0.272 \\
\hline
IP-Adapter SD* & 0.917 & 0.825 & 0.858 & 0.931 & 0.881 & 0.277 & 0.235 & 0.163 & 0.238 & 0.210 \\
\underline{\textbf{+ Conceptrol (Ours)*}} & 0.644 & 0.397 & 0.433 & 0.389 & 0.500 & 0.904 & 0.759 & 0.594 & 0.795 & \underline{\textbf{0.397}} (+89.0\%) \\
\midrule
DreamBooth LoRA SDXL & 0.751 & 0.310 & 0.544 & 0.713 & 0.597 & 0.896 & 0.898 & 0.752 & 0.865 & 0.517 \\
Emu2 SDXL* & 0.665 & 0.551 & 0.447 & 0.443 & 0.526 & 0.729 & 0.730 & 0.558 & 0.691 & 0.364 \\
\hline
IP-Adapter SDXL* & 0.902 & 0.840 & 0.762 & 0.914 & 0.835 & 0.502 & 0.386 & 0.282 & 0.414 & 0.346   \\
\underline{\textbf{+Conceptrol (Ours)*}} & 0.746 & 0.672 & 0.571 & 0.726 & 0.658 & 0.860 & 0.828 & 0.616 & 0.796 & \underline{\textbf{0.524}} (+51.4\%) \\
\midrule
OminiControl FLUX* & 0.555 & 0.182 & 0.476 & 0.306 & 0.438 & 0.945 & 0.920 & 0.821 & 0.910 & 0.398 \\
\underline{\textbf{+Conceptrol (Ours)*}} & 0.642 & 0.344 & 0.612 & 0.344 & 0.556 & 0.907 & 0.866 & 0.781 & 0.866 & \underline{\textbf{0.481}} (+20.9\%) \\

\bottomrule
\end{tabular}
}
\caption{\textbf{Main Results of Dreambench++}. We follow the default evaluation protocol of Dreambench++ to assess the concept preservation and prompt the following scores of various methods, including both fine-tuning approaches and zero-shot adapters (denoted with *). The best results are highlighted with \underline{\textbf{underline}}. Conceptrol significantly enhances the performance of zero-shot adapters, even surpassing fine-tuning methods like Dreambooth across multiple base models.}
\label{tab:main_results}
\end{table*}

\subsection{Evaluation Setup}

\noindent\textbf{Compared methods}. To assess the effectiveness of our method, we systematically compare Conceptrol with other state-of-the-art methods such as Textual Inversion~\cite{gal2022image}, DreamBooth~\cite{peng2024dreambench++}, BLIP Diffusion~\cite{li2024blip} and Emu2~\cite{sun2024generative}. To demonstrate the applicability of Conceptrol across different base models, we apply it with IP-Adapter on UNet-based models including Stable Diffusion 1.5 and SDXL, and OminiControl on FLUX which is DiT-based models.

\noindent\textbf{Evaluation Protocol}. We adhere to the evaluation protocol outlined in DreamBench++~\cite{peng2024dreambench++}, a comprehensive dataset for personalized image generation. This benchmark systematically assesses customization performance in terms of concept preservation and prompt following using a vision-language model GPT4~\cite{openai2024gpt4technicalreport}, demonstrating superior alignment with human preferences compared to other benchmarks~\cite{peng2024dreambench++, gal2022image}. For formal evaluation, personalized generation is formulated as Nash Bargaining problem~\cite{nash1950bargaining}. The target is to maximize the Nash utility, the multiplication of concept preservation, and prompt adherence.

\noindent\textbf{Human Study}. We also conducted a human study using Amazon Mechanical Turk (MTurk)~\cite{mturk} to verify that our method aligns with human preferences. Specifically, participants were presented with pairs of images and asked to select the one that better preserved the original concept and adhered to the prompt. More details are in Appendix.~\red{B}.

\noindent\textbf{Implementation Details}. We use the recommended settings for the compared methods, including guidance scale, number of denoising steps, and conditioning scale, from their original paper or Dreambench++. For our Conceptrol with IP-Adapter on Stable Diffusion 1.5 and SDXL, we use conditioning scale $\lambda$ as 1.0 and conditioning warmup ratio $\epsilon$ as 0.2; with OminiControl on FLUX, we use the conditioning scale as 1.0 and conditioning warmup ratio as 0, where the ablation study of hyperparameter selection is in Sec.~\ref{sec:ablation}.

\begin{figure}
    \centering
    \includegraphics[width=\linewidth]{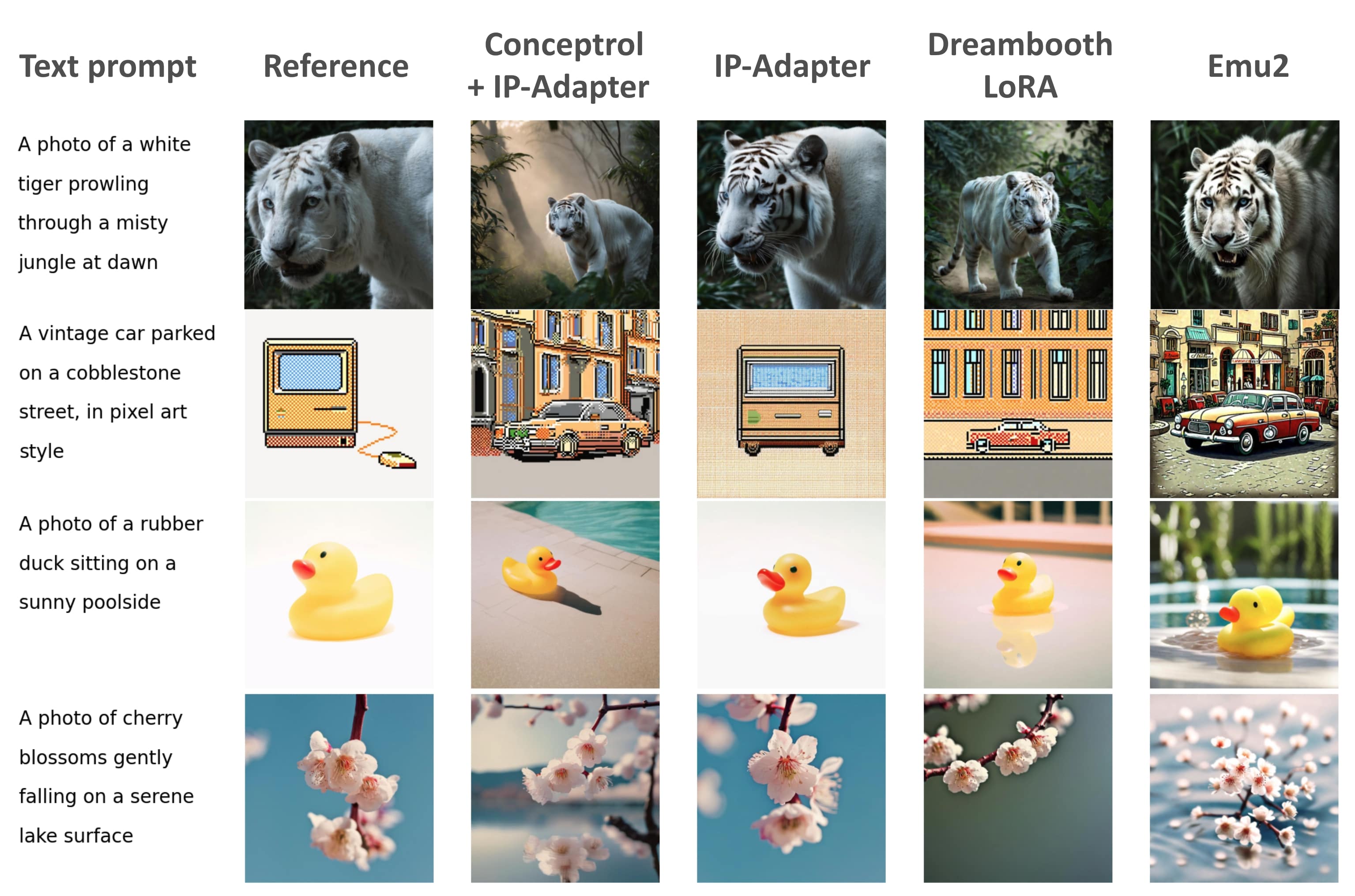}
    \caption{\textbf{Qualitative Results on SDXL.} Compared to the vanilla IP-Adapter, Conceptrol significantly enhances customization across diverse targets and remains competitive with Dreambooth LoRA which requires fine-tuning. We present more qualitative results in Appendix.~\red{B}.}
    \label{fig:exp_vis}
    \vspace{-3px}
\end{figure}

\subsection{Main Results}

We present the main results across different methods, and base models with various personalized targets in Table.~\ref{tab:main_results}.

\noindent\textbf{Freely elicit potential of existing adapters}. With simple control, we can boost the performance of zero-shot adapters on Stable Diffusion 1.5, SDXL, and FLUX by a large margin. Notably, with Conceptrol, the performance zero-shot adapters can even surpass fine-tuning methods such as Dreambooth LoRA (0.397 $>$ 0.359 on Stable Diffusion 1.5, 0.524 $>$ 0.517 on SDXL), indicating the potential of these zero-shot adapters can be further elicited with negligible computation overhead as shown in Fig.~\ref{fig:exp_vis}.

\noindent\textbf{Pareto improvement over human preference}. We report human study results in Fig.~\ref{fig:human_study}. In contrast to the results obtained using GPT-4 evaluation, our method performed similarly to the vanilla IP-Adapter on SD and SDXL in terms of concept preservation, while demonstrating significantly better prompt adherence. This observation is consistent with DreambenchPlus~\cite{peng2024dreambench++}, which reports that human alignment in GPT-4 evaluation is higher for prompt adherence than for concept preservation. Additionally, on FLUX with OminiControl, Conceptrol is able to improve concept preservation without losing prompt following. Overall, the human study results indicate that our method can be regarded as a Pareto improvement, which enhances prompt adherence or concept preservation without sacrificing the other one.

\begin{figure}
    \centering
    \begin{minipage}[b]{0.5\linewidth}
        \centering
        \includegraphics[width=\linewidth]{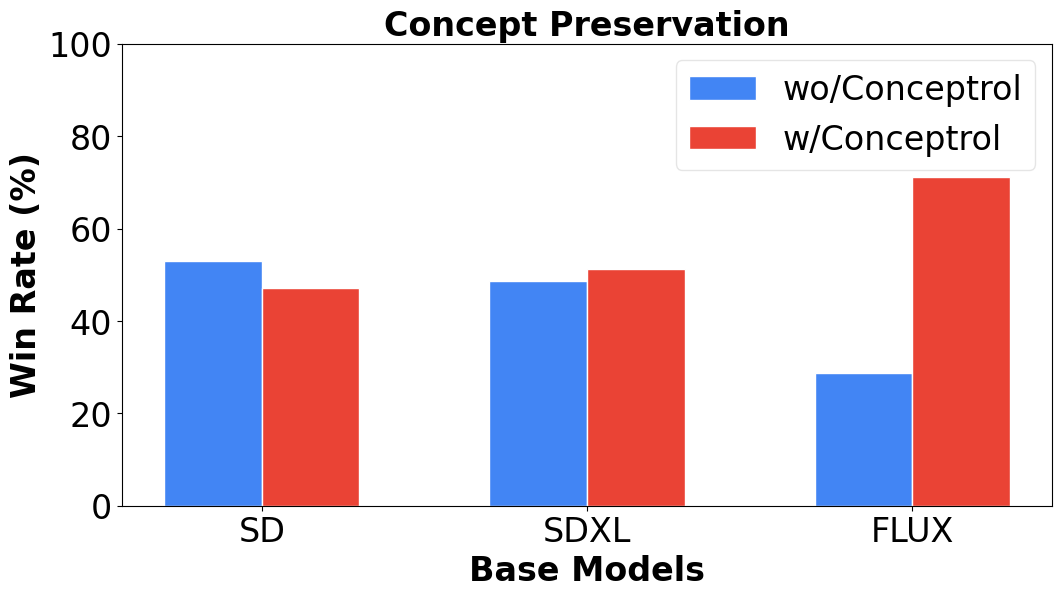} 
    \end{minipage}%
    \begin{minipage}[b]{0.5\linewidth}
        \centering
        \includegraphics[width=\linewidth]{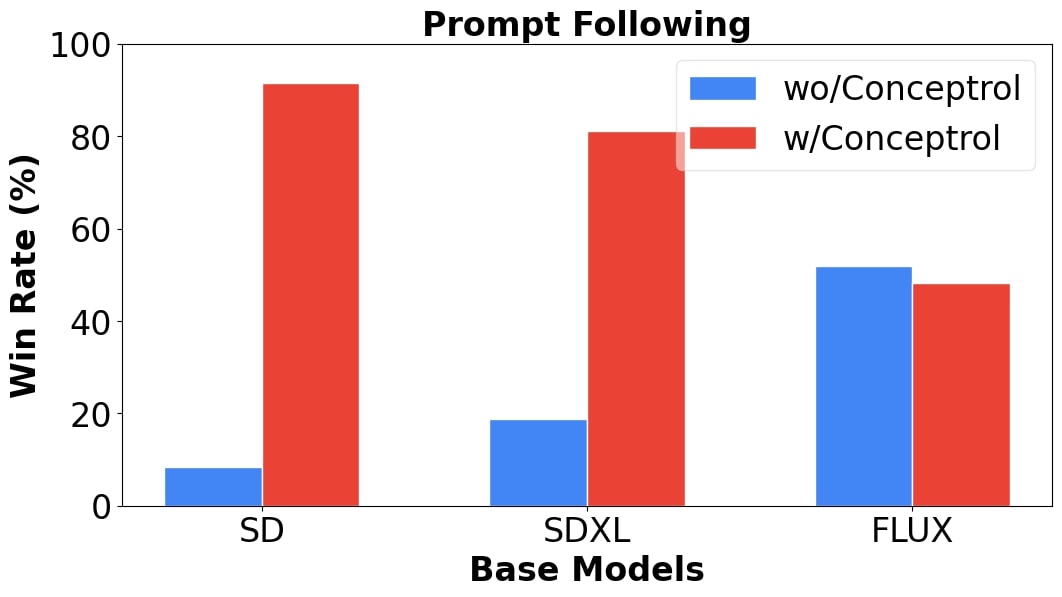}
    \end{minipage}
    \caption{\textbf{Human Study Results}. On SD and SDXL with IP-Adapters, Conceptrol boosts the performance on prompt following while achieving similar performance of concept preservation. On FLUX with OminiControl, Conceptrol can increase the performance on concept preservation with a similar prompt following.}
    \label{fig:human_study}
\end{figure}

\subsection{Ablation Study}
\label{sec:ablation}
We systematically evaluate the impact of each component in our methods including masking mechanism, conditioning scale, and warm-up ratio to the personalization score.

\begin{table}[t]
    \centering
    \scalebox{0.84}{
    \begin{tabular}{c|c|c|c|c}
    \toprule
        Masking method & NFE & CP & PF & CP $\cdot$ PF \\
    \midrule
        non-specific mask & 50 & 0.670 & 0.611 & 0.409 \\
        mask from DOWN.0.0.0 & 50 & 0.611 & 0.582 & 0.356 \\
        \textbf{textual concept mask (Ours)} & 50 & 0.658 & 0.796 & \textbf{0.524} \\
        \hline
        oracle mask (with SAM) & 100 & 0.697 & 0.754 & 0.526 \\
    \bottomrule
    \end{tabular}
    }
    \caption{\textbf{Ablation study on the masking mechanism with SDXL.} Here CP represents the concept preservation score, PF represents the prompt following score, and NFE represents a number of denoising steps. With textual concept mask, we can elicit the potential of a text-to-image model and achieve competitive results with methods using double computation and large segmentation models.}
    \label{tab:mask_ablation}
    \vspace{-5px}
\end{table}

\noindent\textbf{Masking Mechanism}. To evaluate the effectiveness of the textual concept mask, we compare it with three alternative settings: 1) \textit{Non-specific mask}, where the attention mask is directly transferred from the textual concept in each block individually, without using the concept-specific attention block; 2) \textit{Mask from other blocks}, such as DOWN.0.0.0; and 3) \textit{Oracle mask}, where an image is first generated entirely based on the text prompt, followed by segmentation of the subject using SAM~\cite{kirillov2023segment} to extract the mask. As shown in Table~\ref{tab:mask_ablation}, the textual concept mask outperforms the non-specific mask and masks extracted from uninformative attention blocks such as DOWN.0.0.0. Notably, without additional computational overhead or reliance on auxiliary models, the textual concept mask is as competitive as the oracle mask, which requires double the computation and an external large-scale segmentation model.

\begin{figure}
    \centering
    \begin{minipage}[b]{0.45\linewidth}
        \centering
        \includegraphics[width=\linewidth]{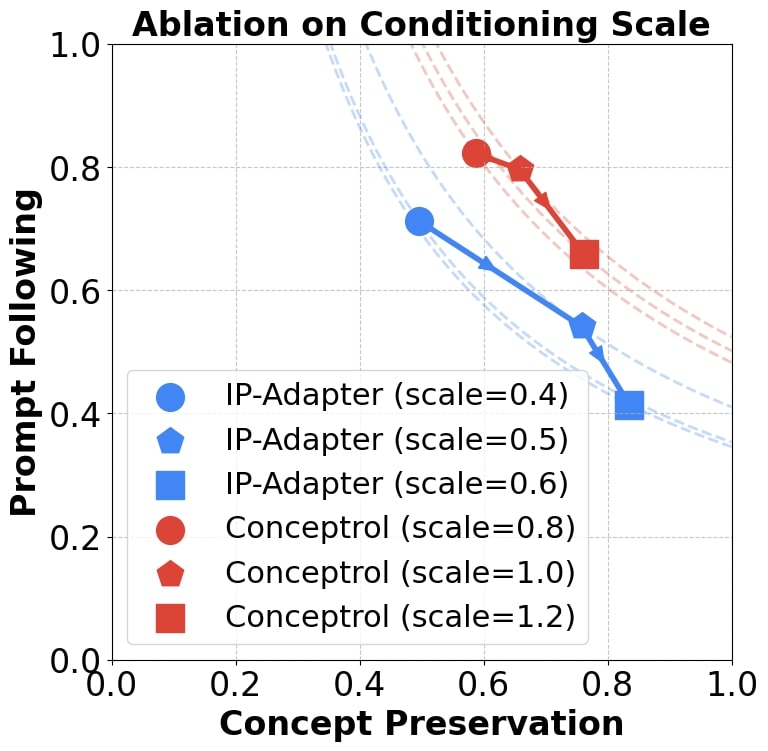} 
    \end{minipage}%
    \hfill
    \begin{minipage}[b]{0.45\linewidth}
        \centering
        \includegraphics[width=\linewidth]{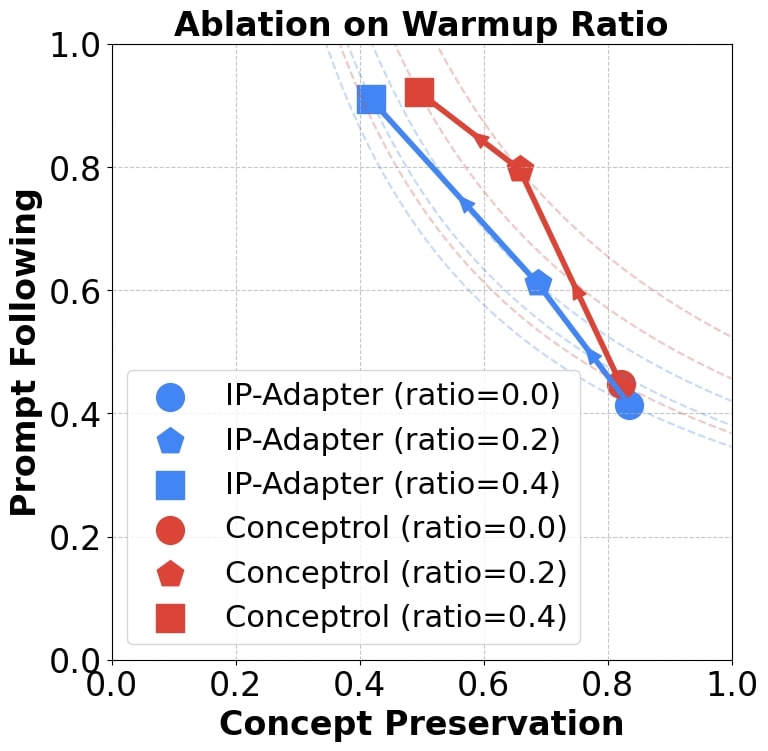}
    \end{minipage}
    \\
    \begin{minipage}[b]{0.5\linewidth}
    \centering
        \small (a)
    \end{minipage}%
    \begin{minipage}[b]{0.5\linewidth}
       \centering
       \small (b)
    \end{minipage}
    \caption{\textbf{Ablation study on (a) conditioning scale and (b) warmup ratio using SDXL.} Each dashed line represents the curve $y = \frac{t}{x}$ across the corresponding data points, where a higher curve in the plot indicates better results for personalization. Conceptrol achieves a superior trade-off curve across both hyperparameters.}
    \label{fig:ablation_others}
\end{figure}

\noindent\textbf{Conditioning Scale}. The conditioning scale defines the default trade-off between concept preservation and prompt adherence in zero-shot adapters. We conduct an ablation study on the original IP-Adapter and its variant under Conceptrol, as shown in Figure~\ref{fig:ablation_others} (a). For both approaches, increasing the conditioning scale enhances concept preservation while diminishing prompt adherence. Notably, Conceptrol achieves a superior trade-off, maintaining a higher multiplied score across different conditioning scale values.

\noindent\textbf{Conditioning Warm-Up Ratio}. This ratio is another important hyperparameter; results are shown in Figure~\ref{fig:ablation_others}(b). The prompt-following score improves as the warmup ratio increases, whereas the concept preservation score decreases. However, Conceptrol consistently improves multiplied scores under each setting. We set the warmup ratio to 0.2 for Conceptrol with IP-Adapter to enhance the prompt following. In terms of OmniControl, we set the warmup ratio to 0.0. Essentially, this is because the textual concept mask of FLUX converges much faster than Stable Diffusion and SDXL. We present more details in Appendix.~\red{C}.
\section{Conclusion}

In this paper, we introduced \textbf{\textit{Conceptrol}}—a simple yet effective plug-and-play method that significantly enhances zero-shot adapters for personalized image generation. Our approach is built on three key observations from our attention analysis: (1) the attention for visual specifications is often misaligned with the customized target; (2) visual specifications can be transferred within regions of high attention; and (3) a textual concept mask can be extracted from specific attention blocks where the target receives high attention. By transferring visual specifications using the textual concept mask, Conceptrol achieves remarkable performance \textit{without additional computation, data, or models}. Our findings underscore the importance of integrating textual concepts into personalized image generation pipelines, even with more data and advanced architectures.

\clearpage

{
    \small
    \bibliographystyle{ieeenat_fullname}
    \bibliography{main}
}

\clearpage

\appendix

\section{Analysis Details}

In our analysis, we investigate whether the attention maps from the base model and the adapter differ, and if they highlight the regions of interest in the generated images.

\begin{figure}
    \centering
    \includegraphics[width=\linewidth]{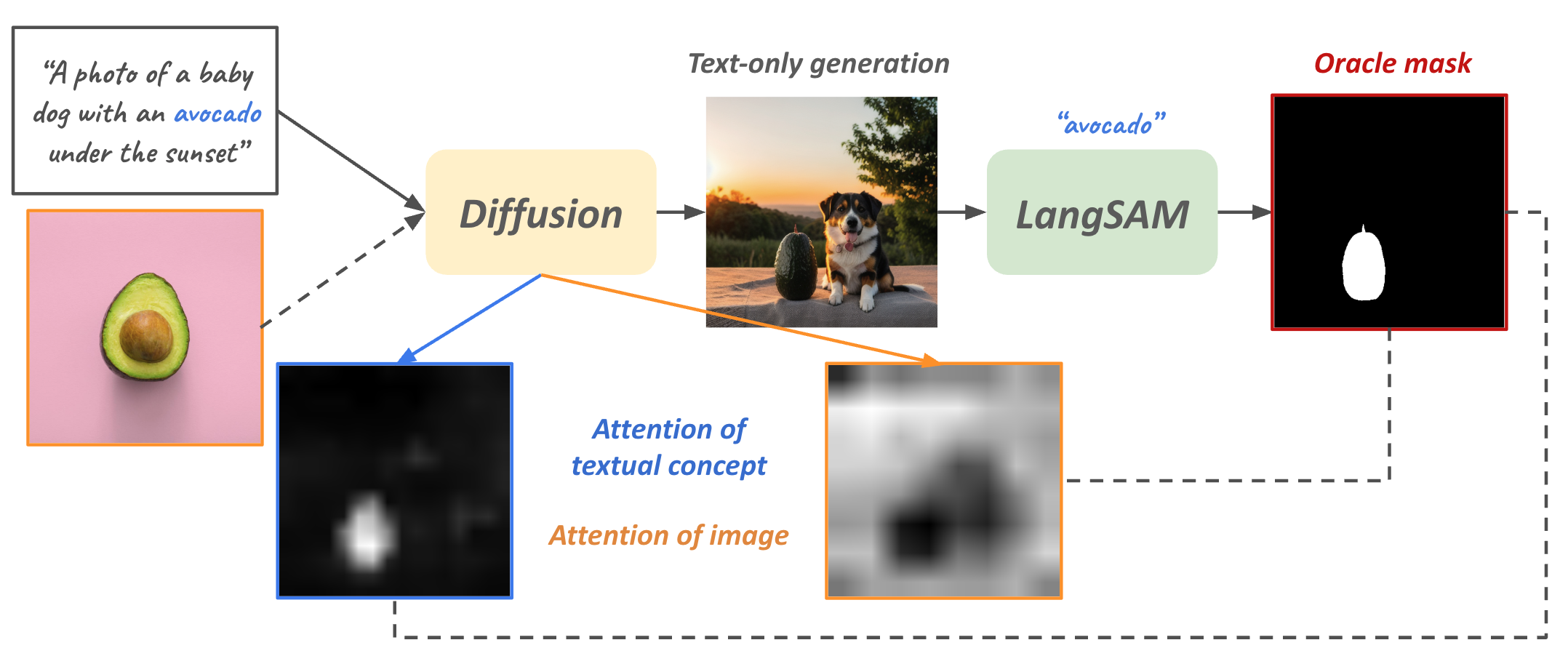}
    \caption{\textbf{Analysis Process.} We generate the results fully based on the text condition. During the generation, we still compute the attention map of reference image bet setting the conditioning scale of reference image as 0 indicated by "$-->$" . After obtaining the text-only results, we use LangSAM~\cite{medeiros2024lang} to retrieve the oracle mask, then compute the AUC between normalized attention map and oracle mask indicated by "$- - -$".}
    \label{fig:attn_process}
\end{figure}

We sample 300 image-text pairs from DreambenchPlus~\cite{peng2024dreambench++} as our analysis targets. The analysis process is shown in Fig.~\ref{fig:attn_process}. For each generated image, we apply LangSAM~\cite{medeiros2024lang} for open-vocabulary segmentation using the provided textual concept $c_\text{concept}$, referring to the resulting segmentation as the oracle mask. We then normalize the attention map obtained during image generation so that its minimum is 0 and its maximum is 1, and compute the AUC between this normalized attention map and the oracle mask. A higher AUC indicates a closer match between the attention map and the oracle mask, suggesting that the attention map accurately identifies the region of interest in the generated image. The following sections detail the analysis experiments. All experiments are conducted over five runs with different random seeds.

\noindent\textbf{Misaligned attention to reference images with textual concept}. As reported in Sec.~\red{3.3}, the highest AUC for the image condition across all blocks averaged on analysis samples, is only 0.38, whereas the highest AUC for the specified textual concept is 0.99 where we also present in Tab.~\ref{tab:analysis_1}. Qualitatively, the attention to image condition is usually globally distributed on every foreground subject, easily leading to artifacts (e.g., a dog and an avocado are rendered as two avocados).

\begin{table}[ht]
    \centering
    \begin{tabular}{c|c|c}
    \toprule
        BLOCK NAME & textual concept & reference image \\
    \midrule
        UP.1.1.0 & 98.89 & 37.72 \\
        UP.1.2.0 & 98.49 & 34.46 \\
        UP.1.0.0 & 99.15 & 24.19 \\
        DOWN.2.1.0 & 95.05 & 15.66 \\
    \bottomrule
    \end{tabular}
    \caption{\textbf{highest AUC of attention on reference image with oracle mask on Stable Diffusion 1.5.} The AUC of textual concept is dominantly higher than the image condition.}
    \label{tab:analysis_1}
\end{table}

\begin{figure*}[ht]
    \centering
    \includegraphics[width=\linewidth]{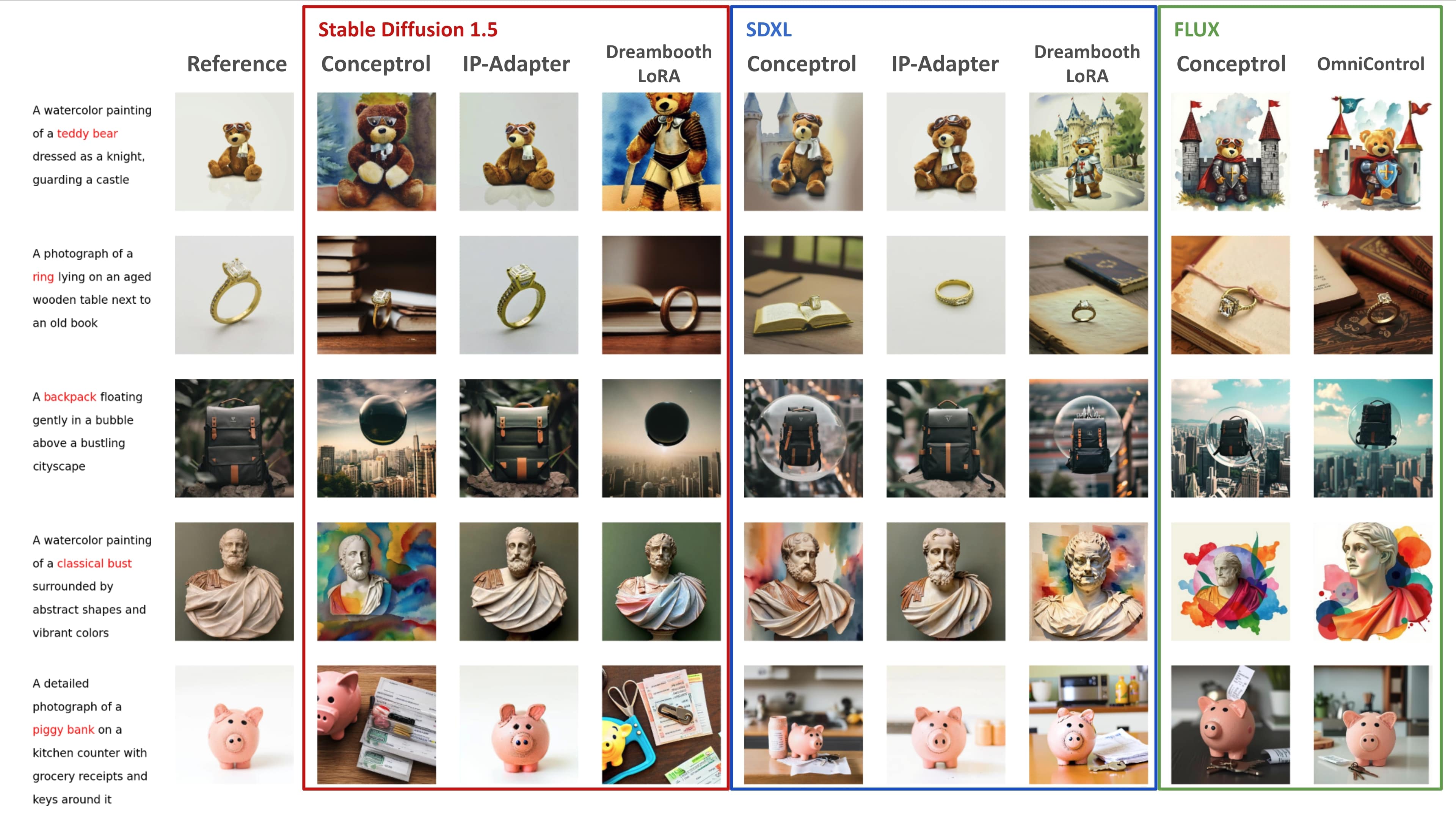}
    \caption{Auxiliary Qualitative Results. Refer to Fig.~\ref{fig:more_1} and Fig.~\ref{fig:more_2} for more results.}
    \label{fig:more_3}
\end{figure*}

\noindent\textbf{Visual specification can be transferred}. We further verify if manually adjusting the attention map of the image condition with the oracle mask can constrain the effect of additional image conditions. The experiment is conducted as follows: 1) Similar to previous analysis, we generate the images fully based on the given text, termed $I_{c_\text{text}}$ and obtain oracle mask $M_{c_\text{concept}}$ from it; 2) We use this oracle mask to mask the attention map of image condition, then generate with text and image together to get another generated image called $I_{c_\text{fused}}$, and obtain its oracle mask similarly $M_{c_\text{fused}}$; 3) We then compute AUC between $M_{c_\text{concept}}$ and $M_{c_\text{fused}}$. If this AUC is higher, it indicates that visual specification is transferred better to the preset region of interest.

In our experiment, the AUC between $M_{c_\text{fused}}$ and $M_{c_\text{concept}}$ is close to 0.99 across each base model and adapters, which strongly show that visual specification can be transferred within regions of high attention score.

\noindent\textbf{Concept-specific blocks for text conditions indicate the region of interest during generation}. In this analysis, we are interested in which attention block of text condition provides the highest AUC with the oracle mask. As reported in Sec.~\red{3.3}, we found that while some blocks provide less information on the region of interest (e.g., the blocks that are closed to the input or output might focus more on the existence of noise), there exist blocks indicating the region of interest with high attention score. In our implementation over Conceptrol, we use \texttt{UP BLOCK 1.0.0} in Stable Diffusion, \texttt{UP BLOCK 0.1.5} in SDXL, and \texttt{BLOCK 18} in FLUX as the concept-specific block to extract the region of interest from the model themselves.

\begin{figure}[bht]
    \centering
    \includegraphics[width=\linewidth]{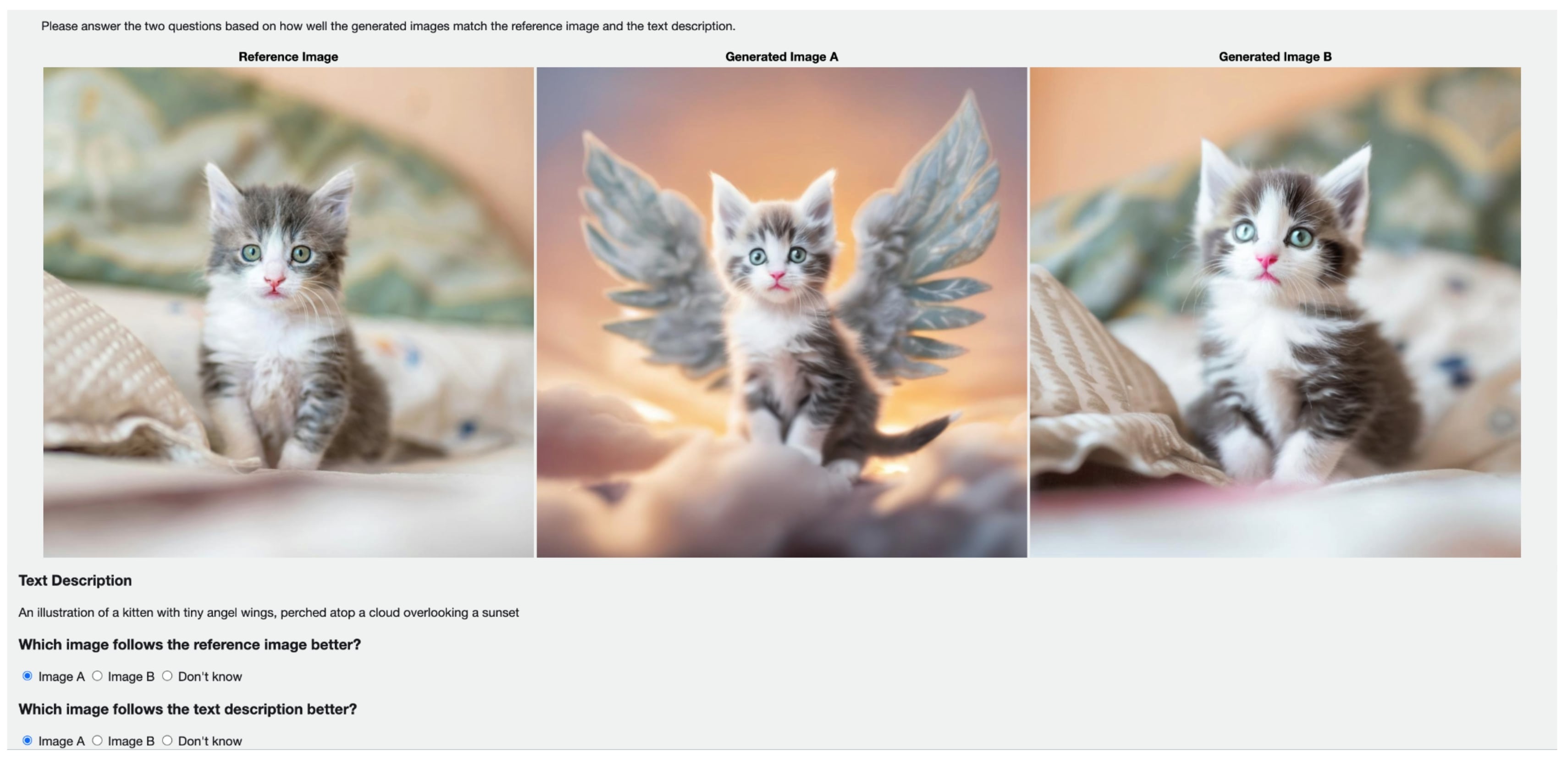}
    \caption{\textbf{Human Study Screenshot}. The user is given a reference image, text prompt, and permuted image pairs generated by vanilla adapters and the version with Conceptrol. They are required to answer two questions: 1) which image preserves the concept better; 2) which image follows the text prompt better?}
    \label{fig:human_study}
\end{figure}

\section{Experiment Details}

\noindent\textbf{Human Study Details.} Figure~\ref{fig:human_study} displays a screenshot of the survey used on Amazon Mechanical Turk. For each base model, we randomly sampled 200 pairs from the results generated by DreambenchPlus~\cite{peng2024dreambench++} and had three human annotators evaluate each comparison. The win rate reported in the main text is computed as follows: in a comparison between method A and method B, if an annotator selects method A, then method A receives a score of 1 and method B receives a score of 0; if method B is selected, the scores are reversed. When an annotator chooses “Don't know,” both methods receive a score of 0.5. After processing all pairs, we calculate the win rate based on the total scores.

\begin{figure*}[ht]
    \centering
    \includegraphics[width=\linewidth]{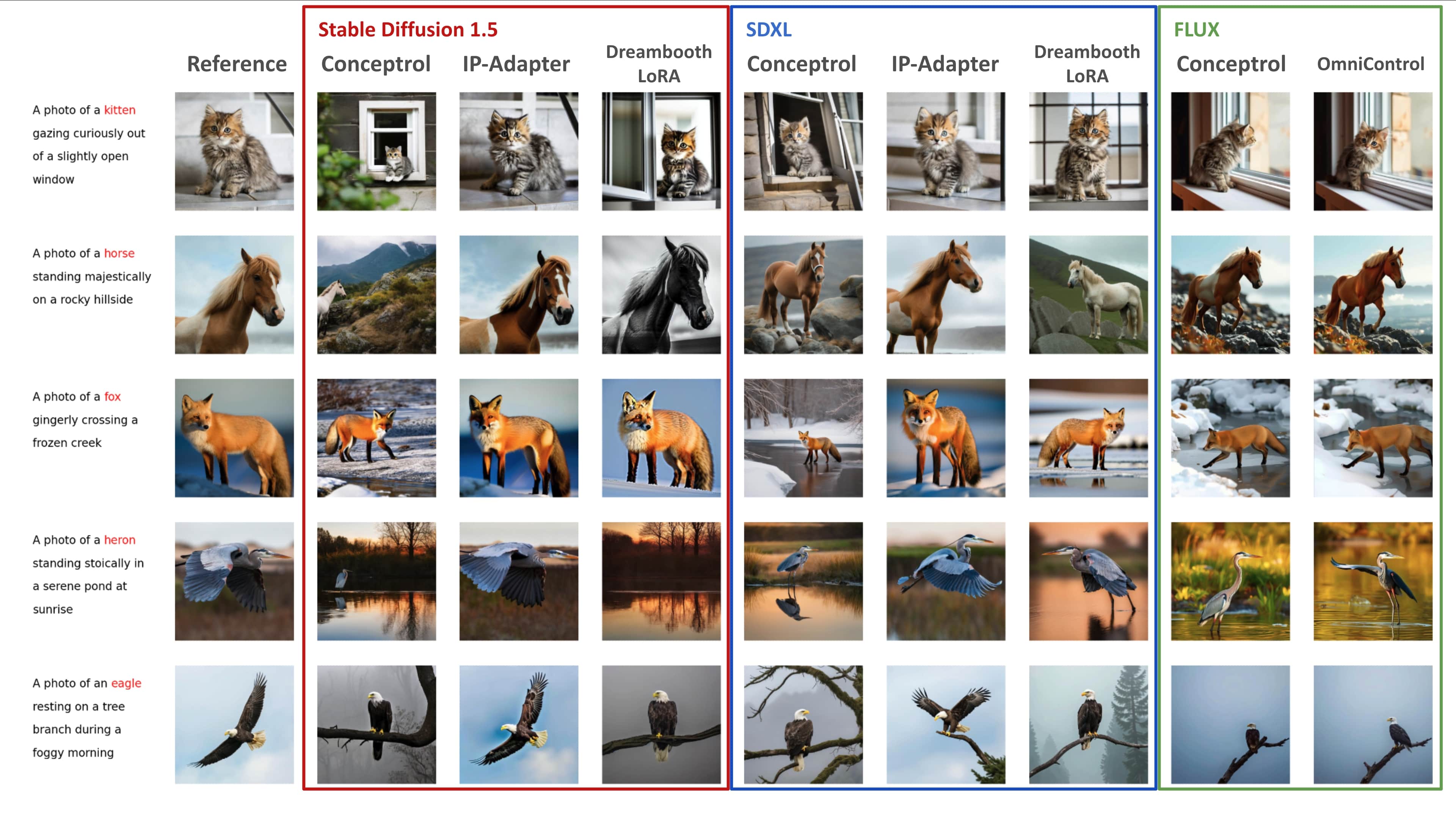}
    \includegraphics[width=\linewidth]{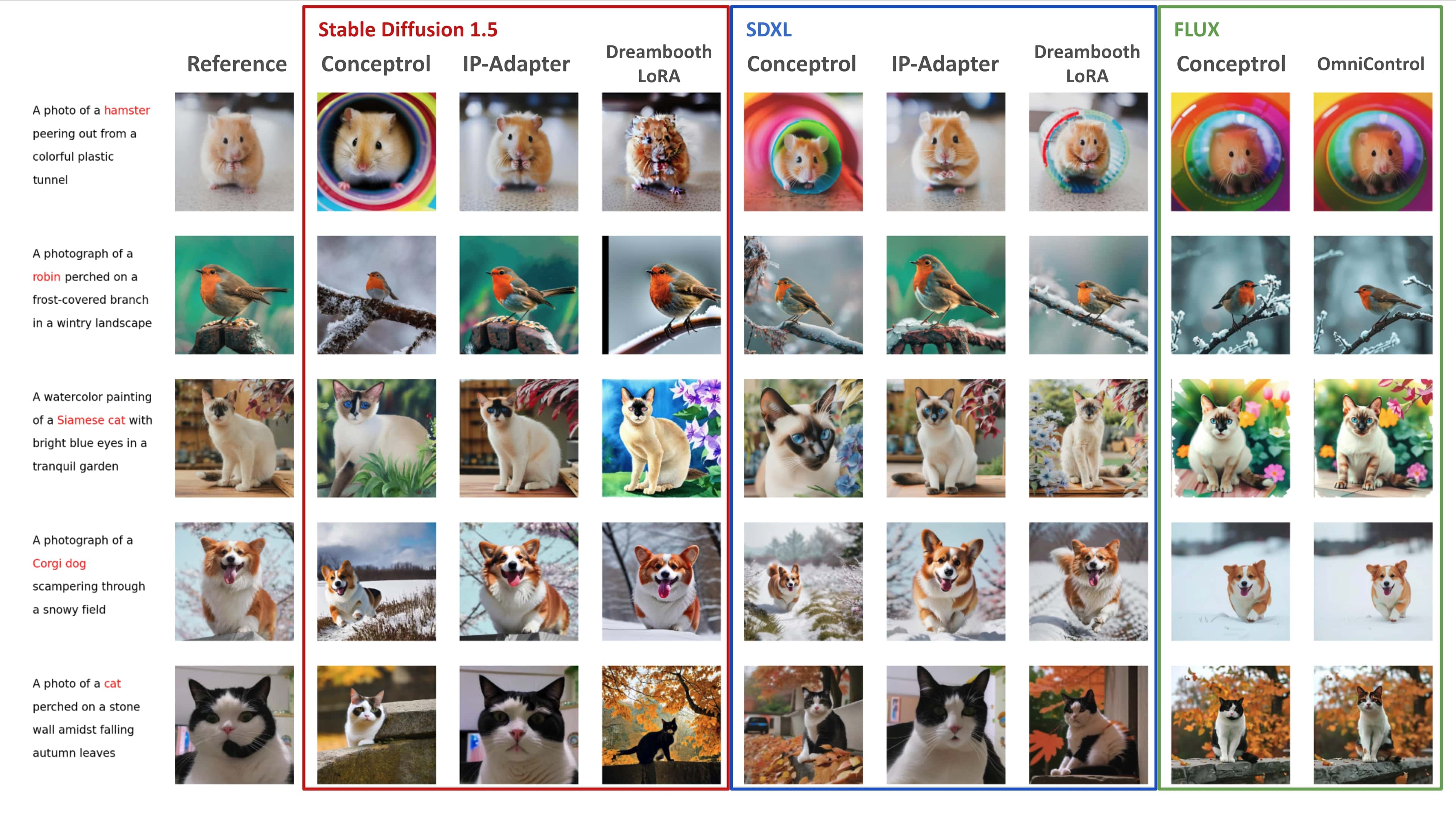}
    \caption{More Qualitative Results.}
    \label{fig:more_1}
\end{figure*}

\begin{figure*}[ht]
    \centering
    \includegraphics[width=\linewidth]{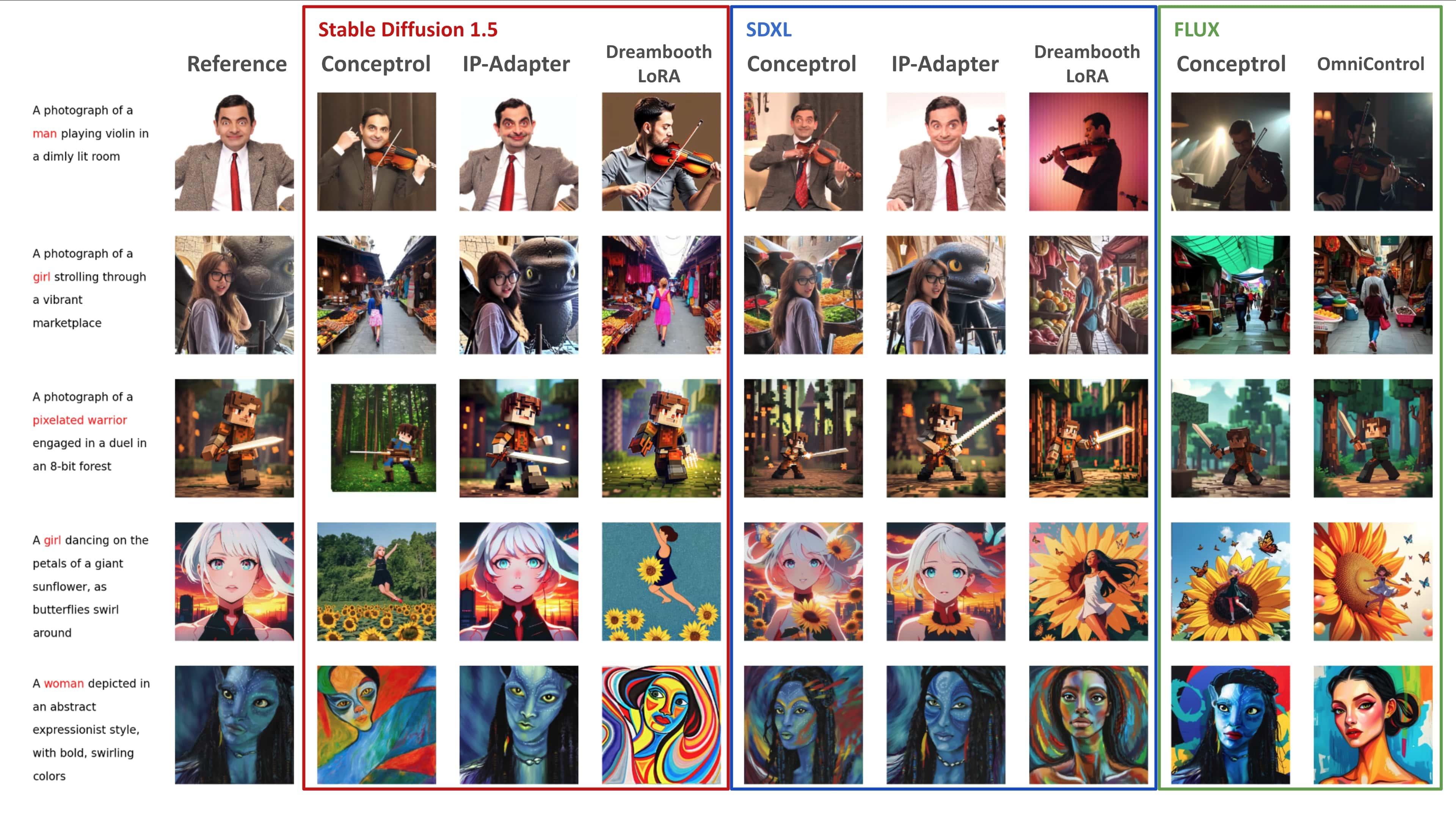}
    \includegraphics[width=\linewidth]{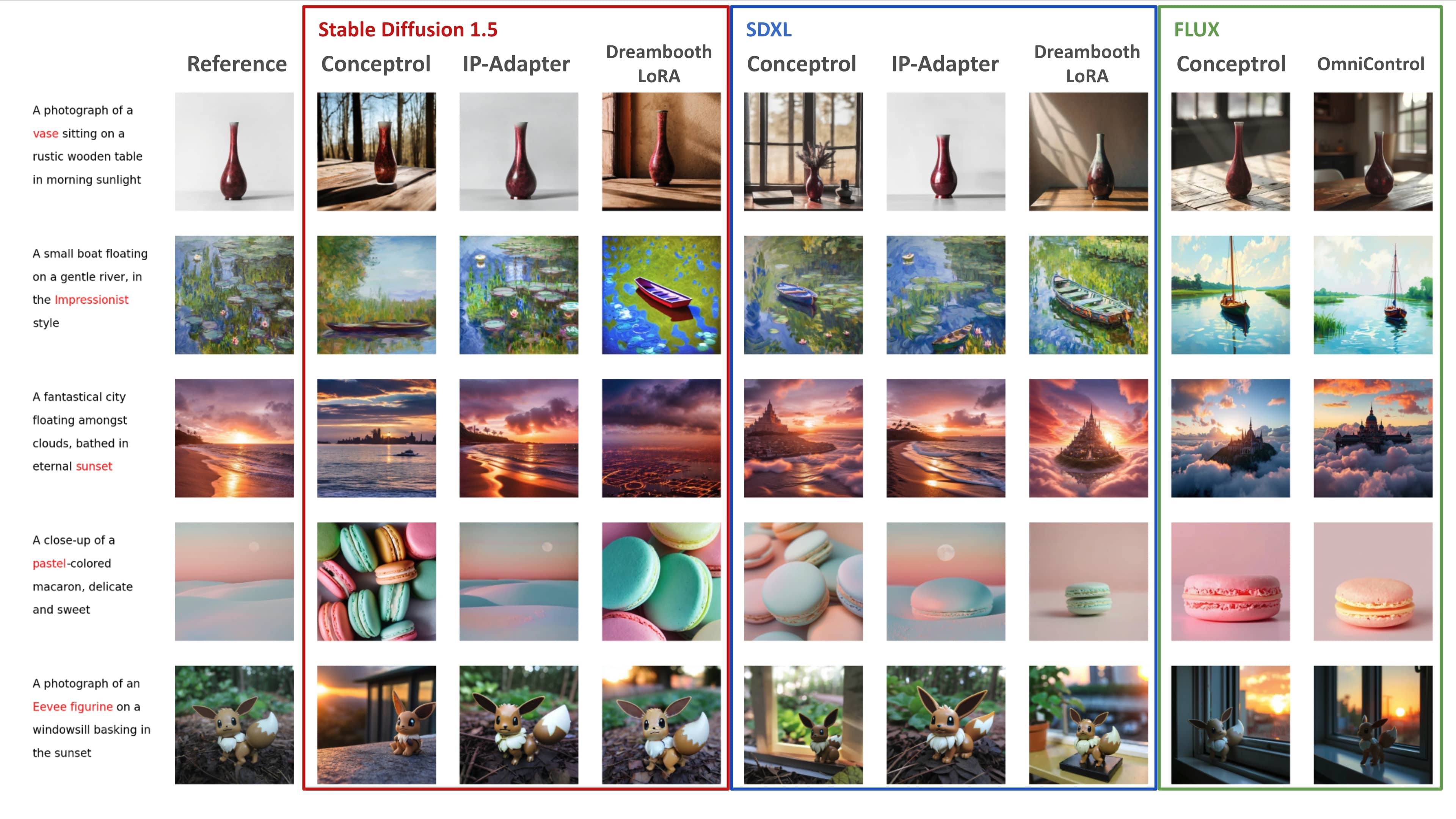}
    \caption{More Qualitative Results.}
    \label{fig:more_2}
\end{figure*}

\noindent\textbf{More Qualitative Results}. We present additional qualitative results in Fig.\ref{fig:more_1}, Fig.\ref{fig:more_2}, and Fig.\ref{fig:more_3} across various customized targets—including animals, humans, objects, and styles. As shown, our method is generally as competitive as Dreambooth LoRA and often outperforms it, without any extra computational overhead and with fewer copy-paste artifacts (e.g., the corgi in the 9th row of Fig.\ref{fig:more_1}). Moreover, our approach improves concept preservation for FLUX when integrated with OmniControl. For instance, in the 10th row of Fig.\ref{fig:more_1}, Conceptrol successfully captures the distinctive features of a cat resembling a telephone operator. Additionally, even when FLUX is trained without human data, Conceptrol enables effective customization, as demonstrated in the 3rd–5th rows of Fig.\ref{fig:more_2} (although several failure cases are observed in the 1st–2nd rows).

\section{Method Details}

\begin{figure}
    \centering
    \includegraphics[width=\linewidth]{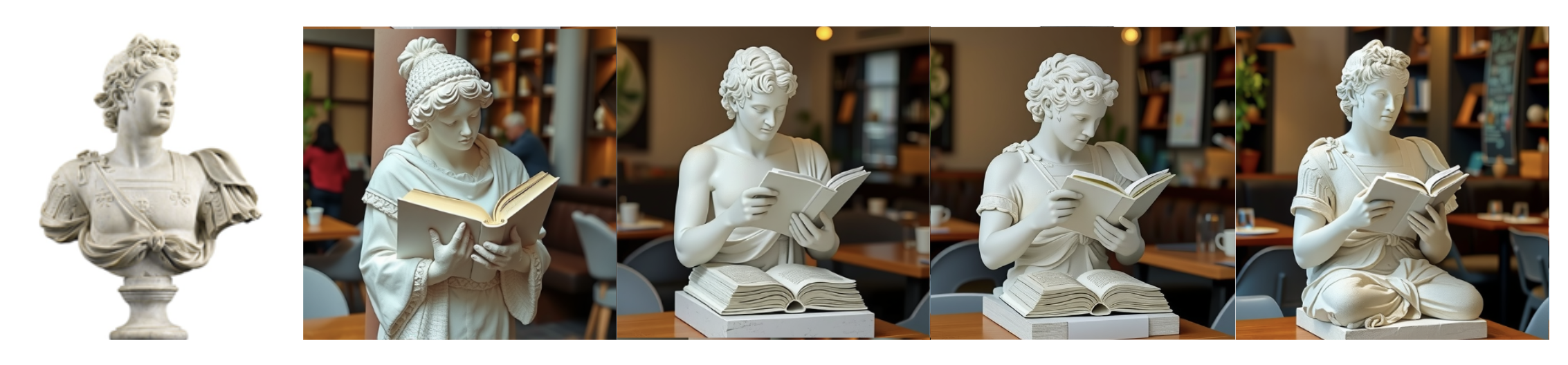} \\
    \begin{minipage}[b]{0.2\linewidth}
        \centering
        (a)
    \end{minipage}%
    \begin{minipage}[b]{0.2\linewidth}
        \centering
        (b)
    \end{minipage}%
    \begin{minipage}[b]{0.2\linewidth}
        \centering
        (c)
    \end{minipage}%
    \begin{minipage}[b]{0.2\linewidth}
        \centering
        (d)
    \end{minipage}%
    \begin{minipage}[b]{0.2\linewidth}
        \centering
        (e)
    \end{minipage}%
    \caption{\textbf{An example of ablation study on Conceptrol with MM-Attention}. All images are generated with the prompt "A statue is reading in the cafe" using FLUX as the base model. From left to right, the images are (a) reference image,  generation (b) without reference image, (c) with OminiControl, (d) with Conceptrol but not suppressing attention of irrelevant text tokens, and (e) with full Conceptrol.}
    \label{fig:ablation_conceptrol_mm}
\end{figure}

\begin{figure}
    \centering
    \includegraphics[width=\linewidth]{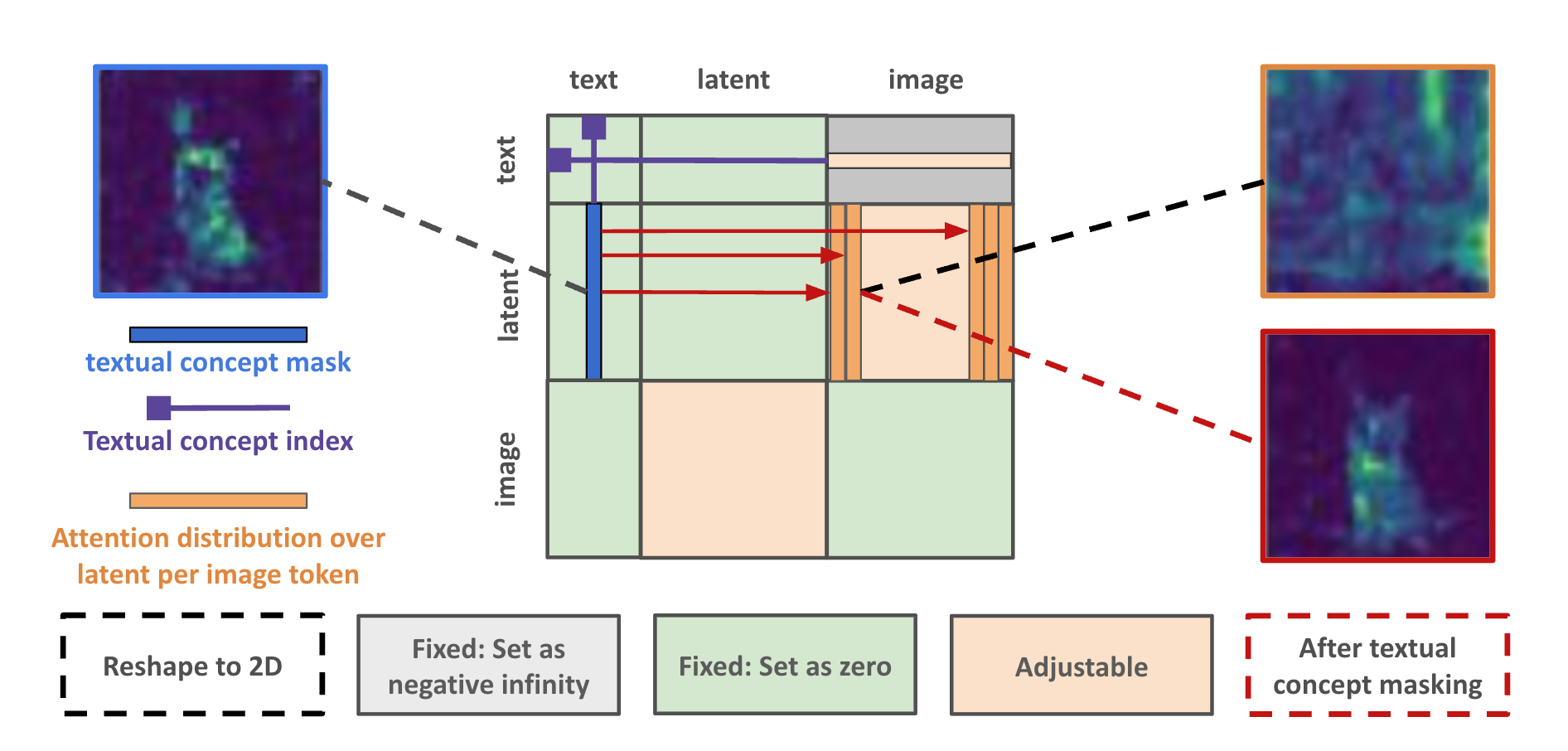}
    \caption{\noindent\textbf{Visualization of adjusted MM-Attention of Conceptrol}. We first locate the textual concept index, i.e., $[i_s, i_e]$, then extract the \blueText{textual concept mask} and apply it to distribution over latent per image token.}
    \label{fig:conceptrol_detail}
\end{figure}

\noindent\textbf{Details of Conceptrol on MM-Attention}. The visualization of the multi-modal attention mask applied in the Conceptrol is shown in Fig.~\ref{fig:conceptrol_detail}. For instance, given the textual prompt "a cat is chasing a butterfly" and a personalized image of a cat, we first index the text token corresponding to "cat" in the prompt, i.e., $[i_s, i_e]$ in the main text. Then we slice the attention of latent to the tokens of textual concept and obtain \blueText{textual concept mask}. Each column vector in the attention of the latent to reference image is subsequently masked with the textual concept mask. Additionally, we filter the attention of irrelevant text tokens to the image. For instance, the text tokens of "chasing" and "butterfly" should not rely on the image token. We present an example in Fig.~\ref{fig:ablation_conceptrol_mm}. Compare (d) with (c), textual concept mask effectively increases concept preservation, i.e., the appearance of the statue is closer to the one presented in the reference image. Compare (e) with (d), suppressing attention of irrelevant text tokens to reference images can reduce the artifact and improve generation quality (e.g. erasing the additional book at the bottom).

\begin{figure}
    \centering
    \includegraphics[width=\linewidth]{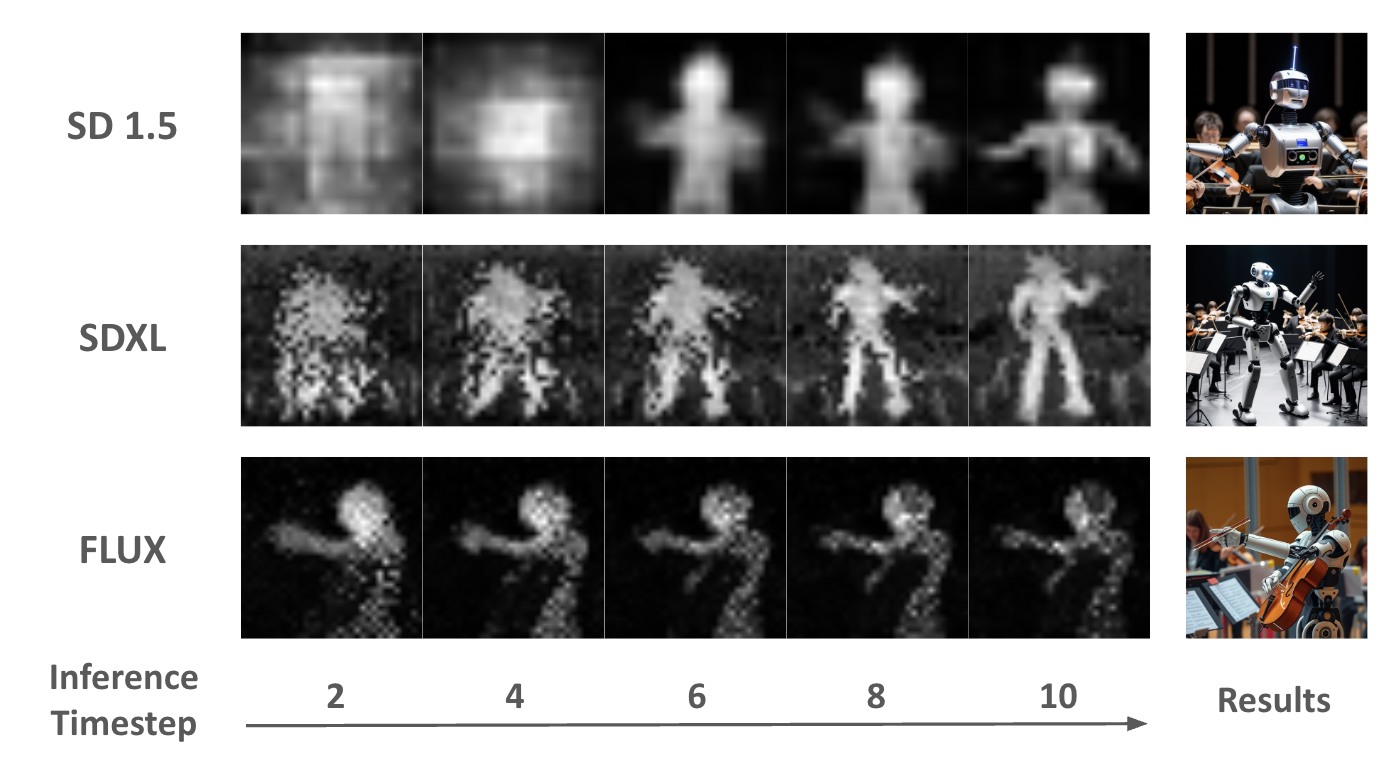}
    \caption{\textbf{Textual Concept Mask of Stable Diffusion, SDXL, and FLUX.} All images were generated using the text prompt “A robot is playing violin,” with “robot” serving as the designated textual concept. As shown, FLUX’s textual concept mask converges more quickly, while the evolution of those in Stable Diffusion and SDXL is comparatively slower.}
    \label{fig:attn_mask_vis}
\end{figure}

\noindent\textbf{Details of Conceptrol Warm-up}. In our experiments, we use a warm-up ratio of 0.2 for Stable Diffusion and SDXL, but 0.0 for FLUX. We observed that the area under the curve (AUC) between the textual concept mask and the ground truth mask is lower for Stable Diffusion and SDXL than for FLUX, as shown in Fig.\red{6} of the main text. In Fig.\red{6}, during the first ten steps, the AUC for Stable Diffusion and SDXL remains below 0.8 for most of the time, whereas FLUX reaches an AUC greater than 0.8 at the very first step. Additionally, Fig.~\ref{fig:attn_mask_vis} visualizes that FLUX’s textual concept mask converges much faster than those of Stable Diffusion and SDXL. Empirically, the AUC of FLUX’s textual concept mask typically converges by the second step (out of fifty steps), while Stable Diffusion and SDXL take about ten steps to converge. Therefore, we apply conditioning warm-up in Stable Diffusion and SDXL, which is not necessary for FLUX.

\end{document}